\pdfoutput=1
\documentclass[11pt]{article}

\usepackage{acl}

\usepackage{times}
\usepackage{latexsym}

\usepackage[T1]{fontenc}
\usepackage[utf8]{inputenc}

\usepackage{microtype}

\usepackage{fancyvrb}
\usepackage{graphicx}
\usepackage{makecell}
\usepackage{mdframed}
\usepackage{subcaption,booktabs}
\usepackage{xspace}
\usepackage{multirow}
\usepackage[hindi,french,german,spanish,main=english]{babel}
\usepackage{wrapfig}
\newcommand{\clasp}[0]{\textsc{Clasp}\xspace}

\newcommand{\claspRS}[0]{\textsc{Clasp-RS}\xspace}
\newcommand{\claspTS}[0]{\textsc{Clasp-TS}\xspace}
\newcommand{\claspGB}[0]{\textsc{Clasp-GB}\xspace}
\newcommand{\claspTB}[0]{\textsc{Clasp-TB}\xspace}

\newcommand{\claspRSGB}[0]{\textsc{Clasp-\{RS,GB\}}\xspace}
\newcommand{\claspTSTB}[0]{\textsc{Clasp-\{TS,TB\}}\xspace}

\newcommand{\pizza}[0]{\textsc{Pizza}\xspace}

\title{\textsc{Clasp}: Few-Shot Cross-Lingual Data Augmentation for\\Semantic Parsing}

\author{Andy Rosenbaum\thanks{~~Corresponding Author} \\
  Amazon, Cambridge, USA \\
  \texttt{andros@amazon.com} \\\And
    Saleh Soltan \\
  Amazon, New York, USA \\
  \texttt{ssoltan@amazon.com}\\\And
  Wael Hamza \\
  Amazon, Dallas, USA \\
  \texttt{waelhamz@amazon.com}\\\AND
  Amir Saffari \\
  Amazon, Cambridge, UK \\
  \texttt{amsafari@amazon.co.uk}\\\And
  Marco Damonte \\
  Amazon, Cambridge, UK \\
  \hspace{0.5cm}\texttt{dammarco@amazon.co.uk}\\\And
  Isabel Groves \\
  Amazon, Cambridge, UK \\
  \hspace{0.5cm}\texttt{isabeg@amazon.co.uk}
  }

\begin{document}
\maketitle
\begin{abstract}
A bottleneck to developing
Semantic Parsing (SP)
models is the need for
a large volume of human-labeled training data.
Given the complexity and cost of human annotation for SP,
labeled data is often scarce, particularly in
multilingual settings.
Large Language Models (LLMs) excel at
SP given only a few examples,
however LLMs are unsuitable for
runtime systems which require low latency.
In this work, we propose \clasp, a simple method to
improve low-resource SP for moderate-sized models:
we generate synthetic data from
\mbox{AlexaTM 20B}
to augment the training set for a model
40x smaller (500M parameters).
We evaluate on two datasets in low-resource settings:
English \textsc{Pizza}, containing either 348 or 16 real examples,
and mTOP cross-lingual zero-shot,
where training data is available only in English, and the model must
generalize to four new languages.
On both datasets, we show significant improvements over strong baseline methods.

\end{abstract}

\section{Introduction and Related Work}

\begin{figure}[h!]
    \centering
    \includegraphics[width=0.45\textwidth]{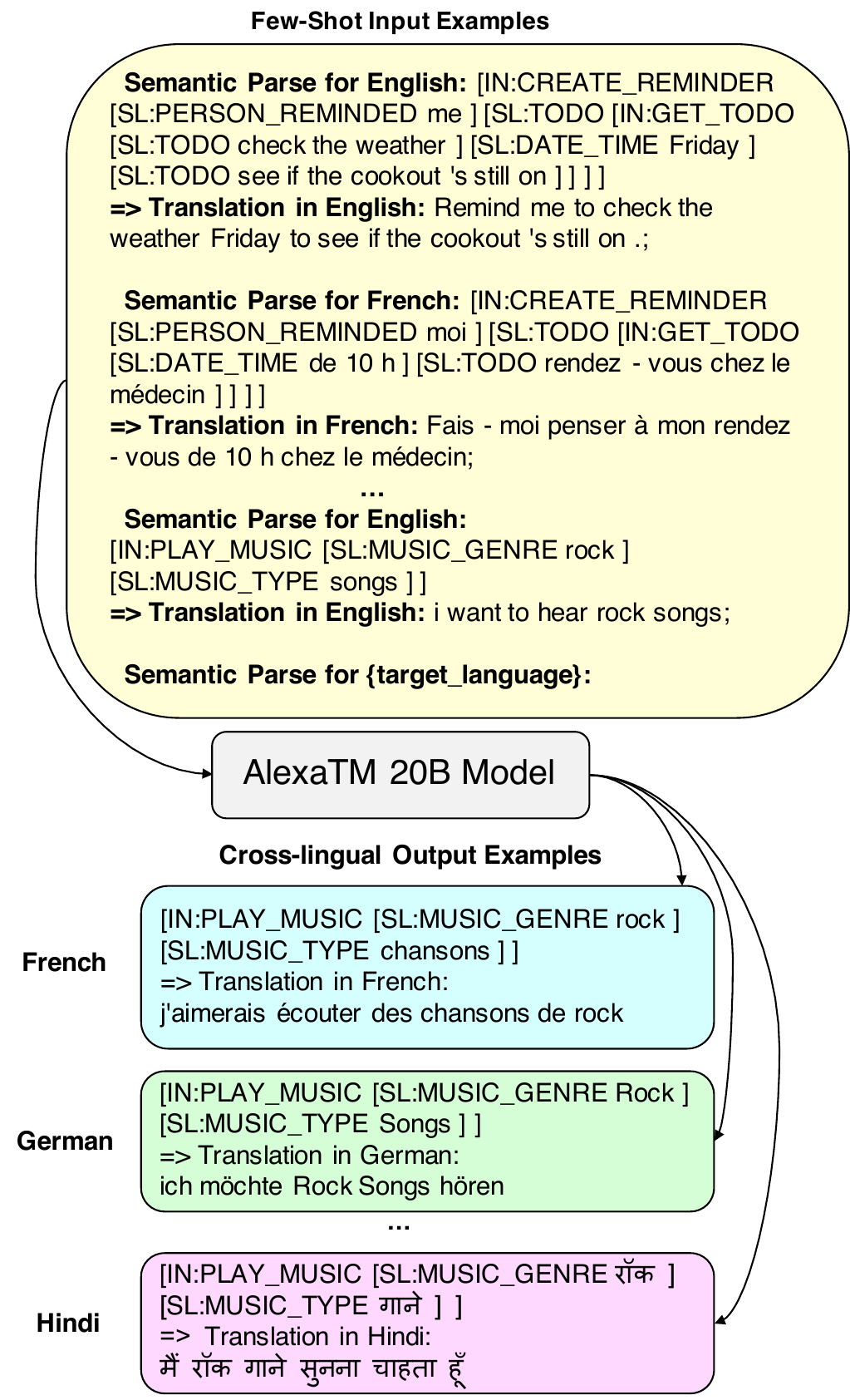}
    \caption{Cross-lingual Data Augmentation: AlexaTM 20B sees only a few examples of mTOP Semantic Parsing and can generate data in multiple languages.}
    \label{fig:main_example_mtop}
\vspace*{-0.5cm}
\end{figure}

\begin{figure*}[h!]
    \centering
    \includegraphics[width=0.9\textwidth]{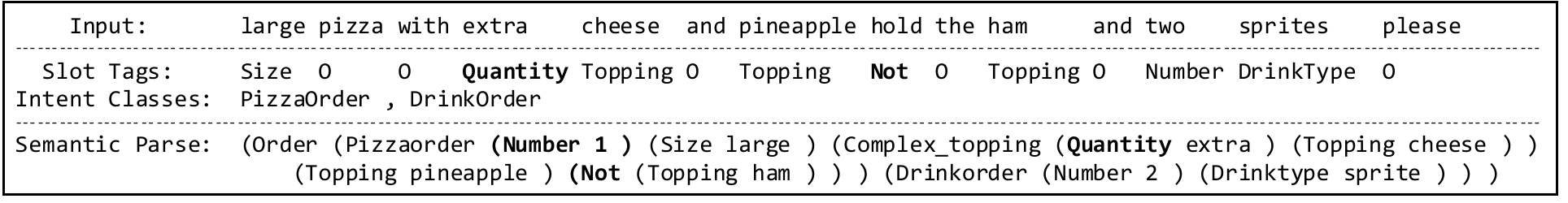}
    \caption{Comparing ``flat'' semantics (Slot Tagging and Intent Classification, upper) to Semantic Parsing (lower).}
    \label{fig:sp_example_pizza}
\end{figure*}

Semantic Parsing (SP) is the task of mapping a natural language sentence to a structured 
representation of its meaning.
SP enables conversational agents
to handle requests
such as ordering pizza, creating reminders, and playing music.
A bottleneck to developing SP models is
their reliance on a large amount of human annotated training data,
which is difficult and expensive to curate (particularly for
multilingual settings)
due to the complexity of the annotation task (Section \ref{sec:motivation}).
While Large Language Models (LLMs) perform well at SP
given limited data \citep{shin-etal-2021-constrained}, they are unsuitable
for runtime systems which require low latency.

Data Augmentation (DA) is a common approach to mitigating data scarcity,
and recently
LLMs are shown to excel at in-context \citep{gpt3}
training data generation for
sentence-level tasks \cite{sahu-etal-2022-data,schick-schutze-2021-generating,Wang2021TowardsZL}.
Fine-tuned LLMs can also generate data for English slot tagging \citep{Lee2021NeuralDA}
and multilingual intent classification and slot tagging \citep{Rosenbaum-et-al-2022-linguist}.
As we discuss in Section \ref{sec:motivation},
SP poses unique challenges for DA,
and remains relatively under-explored in the field.
Prior work
is either limited to heuristic re-combination of the training data 
(\citealt{andreas-2020-good,jia-liang-2016-data})
or else assumes the availability of large-scale unannotated natural data
\citep{yang-etal-2022-addressing}.
Furthermore, there is a gap in the literature
on multilingual DA for SP, 
as most existing work covers only English.

In this work, we extend the general example of DA via LLM prompting to the SP task.
Using AlexaTM 20B \citep{Soltan2022AlexaTM2F}, we generate synthetic training examples for SP,
to augment low-resource settings for moderate-sized models.

We evaluate on two datasets: English \pizza \cite{pizzaDataset} and cross-lingual mTOP \cite{li-etal-2021-mtop}.
On \textbf{\pizza},
we first establish a new SOTA baseline
by improving upon the Canonical Form targets of \citet{rongali2022training}
and tuning the amount of grammar-generated training data,
then show that \textbf{our method improves by 4.79 points (from 80.40 to 85.19)}
on the few-shot n=16 setting on Unordered Exact Match \cite{pizzaDataset}.
On \textbf{mTOP}, we demonstrate \textbf{6.1 points improvement (from 60.3 to 66.4)}
on Exact Match in the cross-lingual one-shot setting, compared to machine translation with slot alignment.

\section{Motivation}
\label{sec:motivation}

\subsection{Why Semantic Parsing?}

Consider an example from \pizza \cite{pizzaDataset}: 
``\textit{large pizza with extra cheese and pineapple hold the ham and two sprites please}''.
As shown in Figure \ref{fig:sp_example_pizza},
SP evolves beyond ``flat'' semantics to exctract complex information
such as the implicit Number slot,
the scope of modifiers Quantity and Not, and the association
between slots and intents.

\subsection{Data Augmentation Challenges for SP}

The core of many standard DA methods is to
modify the text from an existing annotated sample,
assume the same label applies, and accept the novel text-label pair as training data.
For example, a model might paraphrase
``order a pizza with basil''
to ``order a pizza with \textit{extra} basil'',
which would no longer match the original Semantic Parse.

Similarly, in cross-lingual settings
(i.e., data is available in one language 
and the model must perform the task on other languages),
a standard approach for sentence-level tasks
is to translate the text
and keep the label.
For SP however, the target parse must also be updated
with the translated slot values.
\citet{li-etal-2021-mtop}
translate the text then align words to recover the parse.
However, this second alignment step may introduce errors
(Appendix \ref{sec:filtering_mt}).

\section{\clasp Methods}
\label{sec:method}

To address the challenge of
maintaining text-label agreement when generating SP training data,
we propose \clasp (\underline{C}ross-\underline{L}ingual data \underline{A}ugmentation for \underline{S}emantic \underline{P}arsing).
\clasp consists of four methods for prompting LLMs to generate training data, 
either in the Same Language [SL] or Cross-Lingually [CL]:
(1) \textbf{RS}: Replace Slots, Generate Text [SL];
(2) \textbf{TS}: Translate Slots, Generate Text [CL];
(3) \textbf{GB}: Generate Both Parse and Text [SL];
and (4) \textbf{TB}: Translate Both Parse and Text [CL].

\subsection{RS: Replace Slots, Generate Text [SL]}
\label{sec:method_replace_slots}

As shown in Figure \ref{fig:example_pizza_en2} (Appendix \ref{sec:pizza_prompt}),
we start with a real training example, $e_i = (x_i, y_i)$ such as with input text $x_i =$
``\textit{i need to get five small mushroom and bacon pizzas with a pepsi}'', 
and target ground-truth parse $y_i =$
``\textit{(Pizzaorder \ldots{} (Topping \underline{mushroom} ) \ldots{} )}''.
To create a novel training example $e_i' = (x_i', y_i')$
we apply a modification $F(\cdot{})$ on the parse $y_i$ to
obtain $y_i' = F(y_i)$, then prompt a LLM to generate a corresponding text $x_i'$.

Specifically, $F(\cdot{})$
randomly selects one slot (leaf nodes in the parse tree) of $y_i$,
and \textbf{replaces the slot value in the parse} with a different value from a catalog.
In this instance, we replace the Topping ``\textit{mushroom}'' with ``\textit{\underline{spinach}}'',
giving $y_i'$ = ``\textit{(Pizzaorder \ldots{} (Topping \underline{spinach} ) \ldots{} )}''.
To help the model understand how to \textbf{generate the text} $x_i'$,
we include in the prompt 4 other context examples $\{c_j = (x_j, y_j)\}_{j=1}^{4}$
followed by the original example $e_j$, each verbalized as
\texttt{Semantic Parse:} $y_i$ \texttt{Translation in English:} $x_i$.

\subsection{\mbox{TS: Translate Slots, Generate Text [CL]}}

\label{sec:method_translate_slots}

This method extends the idea of \claspRS to
cross-lingual data generation:
we \textbf{translate each slot value} into the target language $l$ and prompt the LLM to \textbf{generate
the corresponding text} in $l$.
See an example in Figure \ref{fig:clasp_ts_mtop} (Appendix \ref{sec:mtop_translate_slot_example}).

\subsection{GB: Generate Both Parse and Text [SL]}
\label{sec:method_gen_both}
\claspRS provides control over the slot values, but cannot
add or remove slot or intents.
Instead, \claspGB \textbf{generates both} the parse and text together,
giving the model flexibility to generate more diverse outputs
(Figure \ref{fig:clasp_gb} in Appendix \ref{sec:example_pizza_gb}).

\subsection{TB: Translate Both Parse and Text [CL]}
\label{sec:method_translate_both}

Given the difficulty of translating a slot value out of context,
which may lead to cascading errors,
we propose to apply the LLM to \textbf{translate both}
the parse and the corresponding text
(Figure \ref{fig:main_example_mtop}).

\section{Experimental Setup}

\subsection{Datasets}
\label{sec:pizza_dataset}

We evaluate \clasp{} on two datasets: \pizza \citep{pizzaDataset}
and mTOP \citep{li-etal-2021-mtop}.

\textbf{\textsc{Pizza}} is a challenging English dataset of SP
for the food ordering domain.
We follow the setting of \citet{rongali2022training}, namely
converting the parse targets to a Canonical Form (CF) closer to natural language; training
on the annotated ``dev'' portion, either full ($n$=348) or
few-shot ($n$=16); and
reporting on the ``test'' portion of 1,357 utterances.
We use 10\% of the test set for checkpoint selection.

We iterate upon the CF targets used for training,
by naturalizing from TOP-style parse to CF while
\textit{preserving the order of sibling slots and intents from the original text}.
(Appendix \ref{sec:data_pre_pizza}). Note that this applies only at training time.\footnote{We release the alternate \pizza dataset used in this paper at \href{https://github.com/amazon-research/pizza-semantic-parsing-dataset/tree/main/data/alternate-canonical}{https://github.com/amazon-research/pizza-semantic-parsing-dataset/tree/main/data/alternate-canonical}.}

\pizza also provides 2.5M grammar-generated ``train'' samples,
and catalogs of values for each slot.

\textbf{mTOP} \citep{li-etal-2021-mtop} is a larger-scale multi-lingual
SP dataset covering 11 domains and 6 languages.
The splits are ``train'' (15,667 English, 10k-11k others),
``validation'' (2,235 English, 1k-2k others), 
and ``test'' (4,386 English, 2k-3k others).
We follow a cross-lingual one-shot setting:
full training and validation data is available for English only,
we use one training example from each other language
for in-context prompts (Appendix \ref{sec:prompt_utts}),
and we test on all languages, however excluding Thai which
is not supported by our pre-trained LMs.
The mTOP dataset
provides two options for the input text, either ``Utterance'',
or ``tokens''.
We use space-joined ``tokens'' which resolves many (although not all)
string matching anomalies (Appendix \ref{sec:data_pre_mtop}).

\subsection{Baselines}
\label{sec:baselines}
For \textbf{\pizza}, we cite \citet{rongali2022training}, who
fine-tune BART 
\citep{lewis-etal-2020-bart}, including joint training with
auxiliary tasks and constrained decoding.
We also explore using various amounts $m$ of (grammar-generated) train data, both in isolation
and mixed with the (annotated) dev set.
Selecting the best-performing $m$ from values between 348 and 174,000
(Appendix \ref{sec:learning_curves})
we use $m$=69,600 for train in isolation.
For combining with dev $n$=348 / $n$=16,
we use train $m$=3,480 / $m$=104,400.
For combining with dev and \clasp, we always use train $m$=348.

For \textbf{mTOP}, we implement machine translation of the text,
via Opus MT \citep{tiedemann-thottingal-2020-opus} and
via the 20B model (using a one-shot in-context prompt, Figure \ref{fig:example_in_context_text_mt} in Appendix \ref{sec:example_in_context_mt_sent}).
We use Sim-Align \citep{jalili-sabet-etal-2020-simalign} (Appendix \ref{sec:simalign_settings}) to align the
translated sentence to the original English, to recover the target-language parse.

\subsection{\clasp Settings}
For \textbf{\pizza}, 
we apply two \clasp methods:
\claspRS (Sec. \ref{sec:method_replace_slots}) and
\claspGB (Sec. \ref{sec:method_gen_both}) to generate novel
training data based on the dev set.
For each method, we generate $k$=3,480 samples
We also try including the union of data from the two \clasp methods,
referred to as \claspRSGB.

For \textbf{mTOP}, we use \claspTS (Sec. \ref{sec:method_translate_slots}) and
\claspTB (Sec. \ref{sec:method_translate_both})
to generate training data in other languages from the English source.
We select a single example from each of the four target languages
(de, es, fr, and hi; shown in Appendix \ref{sec:prompt_utts})
to use in one-shot prompts for generation. We filter the outputs as described in Appendix \ref{sec:filtering}.

Regardless of which and how much data we add, we always
up-sample the non-synthetic data source (dev for \pizza, English data for mTOP) to account for 50\% of the mass
of utterances seen during training, and scale down the number of epochs
to fix the total number of \textit{model updates} across experiments.

\subsection{Metrics}
We use the form of Exact Match (EM) standard for each dataset:
Unordered Exact Match (UEM) \citep{pizzaDataset} for \pizza,
which is invariant to different
order of sibling nodes in parses;
and Space- and Case-Insensitive Exact Match (SCIEM) (Appendix \ref{sec:sciem_metric}) for mTOP,
which is invariant to different spacing and casing of slot values.

\subsection{Models}
\label{sec:models}

For \clasp data generation, we leverage in-context learning with
AlexaTM 20B \citep{Soltan2022AlexaTM2F}.

For Semantic Parsing fine-tuning (\citet{Rongali2020DontPG}, details in Appendix \ref{sec:hyperparams}), we use \mbox{AlexaTM-Large 500M}, a 500-million-parameter
seq2seq
Transformer \citep{Vaswani-et-al-2017-Transformer}
pre-trained similarly to AlexaTM 20B \citep{Soltan2022AlexaTM2F},
however with denoising objective only (no Causal Language Modeling).
This model has 12 encoder, 12 decoder layers, and 1024 hidden size (same as (m)BART \citep{liu-etal-2020-multilingual-denoising}).
For mTOP we use sentinel words \cite{raman-etal-22-st-s2s}
which function similarly to pointers (Appendix \ref{sec:sentinel_words_mtop}).
At test-time inference, we use the top-1 hypothesis from beam search 4 (Appendix \ref{sec:ablation_beam}).

\section{Results}

\subsection{\pizza Results}
\label{sec:pizza_results}

Results are presented in Table \ref{tab:pizza_results}.
We first note that applying our Fixed Canonical Form to dev-only provides
a very large boost in performance, from 82.54/21.00 for $n$=348/$n$=16 to
90.05/58.00, an improvement of 7.51 and 37.00 points, respectively.
For $n$=16, dev-only with Fixed CF already
outperforms the best system reported by \citeauthor{rongali2022training}.
We show that training data (which is grammar-generated) on its own
under-performs at 59.84, however it can help a lot
when combined with the dev set, providing 92.70/80.40.

Both \clasp methods improve significantly over dev-only:
\claspRS provides 92.04/60.65 and \claspGB
provides 93.52/77.75.
Combining data from the \clasp methods (\claspRSGB)
shows a slight improvement
on n=348, however is 2.14 points behind \claspGB alone on n=16.
Finally, our best performing system
uses the fixed Canonincal Form with
data from dev, train, and both \clasp methods together,
obtaining a new SOTA by a wide margin: \textbf{95.06 for n=348} setting, and \textbf{85.19 for n=16} setting.

\begin{table}[h!]
\footnotesize
\centering
\begin{tabular}{lcc}
\hline
 Data                   &   \multicolumn{2}{c}{Unordered EM} \\
 \hline
 \multicolumn{1}{c}{Original CF} & $n$=348 & $n$=16 \\
 \hline
 dev-only (ours)     & 82.54 & 21.00 \\
 \hline
 dev-only (\citeauthor{rongali2022training})       & 87.25 & 16.95 \\
 \citeauthor{rongali2022training} best       & -- & 49.89 \\
 \hline
 \multicolumn{2}{c}{Fixed CF (all ours)} \\
 \hline
 dev-only               & 90.05 & 58.00 \\
 train-only             & 59.84 & 59.84 \\
 dev+train              & 92.70 & \underline{80.40}\\
\hline
 dev+\claspRS   & 92.04 & 60.65\\
 dev+\claspGB & 93.52 & 77.75\\
 \hline
 dev+\claspRSGB & \underline{93.81} & 75.61 \\
 dev+train+\claspRSGB        & \textbf{95.06} & \textbf{85.19}\\
\hline
\end{tabular}
\caption{Results on \pizza dataset with Unordered Exact Match (UEM) metric.
The best and second-best numbers are bolded and underlined, respectively.
Original CF is the Canonical Form of \citet{rongali2022training}.
Fixed CF is our fixed Canonical Form (Sec. \ref{sec:pizza_dataset}), and
$n$ is the number of samples available from the dev set.}
\label{tab:pizza_results}
\end{table}

\subsection{mTOP Results}
\label{sec:mtop_results}

Results are presented in Table \ref{tab:mtop_table},
where the main focus is on ``avg-0s'' (``average-zero-shot''),
the average across the non-English languages.
Training on English data only (``en-only'') is a lower bound of 45.3,
and training on all languages together (``ALL'') is an upper bound of 73.5,
i.e. a gap of 28.2 points.
The baseline MT with Slot-Alignment (``MT-Opus'') provides 15.0 points improvement over ``en-only'', from 45.3 to 60.3. Scaling up the MT model size (``MT-20B'')
does not provide improvement, matching ``MT-Opus'' at 60.3.

\begin{table}[h!]
\setlength{\tabcolsep}{5.0pt}
\centering
\footnotesize
\begin{tabular}{l| c | cccc | c}
\hline
 \makecell{Non-en \\ data}             &   en &   de &   es &   fr &   hi &   \makecell{avg\\0s} \\
\hline
\multicolumn{7}{c}{Lower/Upper Bounds and Baseline} \\
\hline
 en-only                & 83.1 & 47.3 & 51.0 & 54.8 & 28.2 &     45.3 \\
 ALL                    & 83.3 & 70.3 & 77.3 & 75.9 & 70.5 &     73.5 \\
 MT-Opus          & 83.0 & 63.8 & 65.0 & 65.1 & 47.4 &     60.3 \\
 \hline
\multicolumn{7}{c}{Single Methods} \\
\hline
 MT-20B           & 83.3 & 63.8 & 64.3 & 65.2 & 47.8 &     60.3 \\
 \claspTS               & 82.9 & 62.8 & 62.6 & 67.2 & 57.9 &     62.6 \\
 \claspTB               & 83.3 & 65.4 & 64.4 & 66.3 & 54.7 &     62.7 \\
 \hline
\multicolumn{7}{c}{Combination of Methods} \\
 \hline
 \makecell[l]{\clasp-\\~~\{TS,TB\}}          & 83.4 & 64.2 & 63.7 & 68.4 & \underline{59.2} &     63.9 \\
 \hline
 \makecell[l]{\clasp-\\~~\{TS,TB\}\\+MT-20B} & \underline{83.8} & \underline{66.3} & \underline{65.9} & \underline{69.0} & \textbf{59.7} &     \underline{65.2} \\
 \hline
 \makecell[l]{\clasp-\\~~\{TS,TB\}\\+MT-20B\\+MT-Opus} & \textbf{84.4} & \textbf{66.7} & \textbf{68.1} & \textbf{72.6} & 58.1 &     \textbf{66.4} \\
\hline
\hline
\multicolumn{7}{c}{CRISS with Pointers \citep{li-etal-2021-mtop} (for reference only)} \\
\hline
 en-only   &   84.2 &   36.1 &   48.6 &   46.6 &   31.2 &   40.6 \\
  ALL       &   84.1 &   74.4 &   79.1 &   77.7 &   74.7 &   76.5 \\
 MT  &   84.2 &   62.8 &   73.3 &   71.7 &   63.2 &   67.8 \\
\hline
\end{tabular}
\caption{Our mTOP results, where `avg-0s' is averaged across the non-en languages.
\citet{li-etal-2021-mtop} is cited for reference only,
and are not directly comparable due to using a stronger backbone model
(CRISS, \citep{tran-etal-2020-criss}) with a higher upper bound (``ALL'').
Our best result is bolded, and our second best is underlined.}
\label{tab:mtop_table}
\end{table}

\claspTS and \claspTB provide 62.6 and 62.7, respectively,
while their combination (\claspTSTB) improves further to 63.9.
Adding data from ``MT-20B'' increases to 65.2,
and finally by combining data from both \clasp methods
and both MT models, our best result is 66.4, i.e.
\textbf{6.1 points improvement} over the baseline.
The gain is particularly large for
\textbf{Hindi: 12.3 points improvement}  over the baseline
(from 47.4 to 59.7).

%\newpage
\section{Conclusion and Future Work}

We have demonstrated \clasp, a simple method to generate synthetic
training data for
multi-lingual Semantic Parsing by prompting a frozen
Large Language Model.
In very low-resource (n=16 and n=1) settings, on two datasets covering five languages,
we show significant improvements over strong baseline methods.
In future work, we would like to
evaluate on more languages and datasets,
combine our method with CRISS style pre-training,
and extend our method to more tasks such as Text-to-SQL and Code Generation.

\bibliography{_main}
\bibliographystyle{acl_natbib}

\onecolumn
\appendix
% \pagenumbering{arabic}
% \renewcommand*{\thepage}{A\arabic{page}} 
% \setcounter{figure}{0}    
% \renewcommand*{\thefigure}{A\arabic{figure}} 

% \section*{Acknowledgements}

\section{Sample Model outputs}

\subsection{Example of \claspRS: Replace Slots and Generate Text}
We show an example of \claspRS (Replace Slots and Generate Text) in Figure \ref{fig:example_pizza_en2}.
\label{sec:pizza_prompt}

\begin{figure}[h!]
\begin{Verbatim}[fontsize=\scriptsize, frame=single, commandchars=\\\{\}]
INPUT:
[CLM] \textcolor{red}{Semantic Parse:} (Order 
  (Pizzaorder (Number a ) (Size medium ) (Style supreme ) ) 
  (Drinkorder (Number a ) (Drinktype sprite ) ) );
\textcolor{teal}{Translation in English:}
  order me a medium supreme pizza and a sprite;
\textcolor{red}{Semantic Parse:} (Order 
  (Pizzaorder (Number two ) (Topping bacon ) (Topping onion ) ) 
  (Drinkorder (Number a ) (Size large ) (Drinktype mountain dew ) ) );
\textcolor{teal}{Translation in English:}
  put in my order for two bacon and onion pizzas and include a large mountain dew;
\textcolor{red}{Semantic Parse:} (Order 
  (Pizzaorder (Number two ) (Size large ) (Topping pepperoni ) (Topping mushrooms ) ) 
  (Drinkorder (Number four ) (Size large ) (Drinktype cherry cokes ) ) );
\textcolor{teal}{Translation in English:}
  two large pizzas with pepperoni and mushrooms and four large cherry cokes;
\textcolor{red}{Semantic Parse:} (Order 
  (Pizzaorder (Number one ) (Size small ) (Topping yellow peppers ) (Topping olives ) ) 
  (Drinkorder (Number two ) (Containertype cans ) (Drinktype coke ) ) );
\textcolor{teal}{Translation in English:}
  place an order for one small pizza with yellow peppers and olives 
  and also include two cans of coke with it;
\textcolor{red}{Semantic Parse:} (Order 
  (Pizzaorder (Number five ) (Size small ) (Topping \underline{mushroom} ) (Topping bacon ) ) 
  (Drinkorder (Number a ) (Drinktype pepsi ) ) );
\textcolor{teal}{Translation in English:}
  i need to get five small \underline{mushroom} and bacon pizzas with a pepsi;
\textcolor{red}{Semantic Parse:} (Order 
  (Pizzaorder (Number five ) (Size small ) (Topping \textbf{spinach} ) (Topping bacon ) ) 
  (Drinkorder (Number a ) (Drinktype pepsi ) ) );
\textcolor{teal}{Translation in English:}

OUTPUTS:

0: five small \textbf{spinach} and bacon pizzas with a pepsi

1: put in my order for five small \textbf{spinach} and bacon pizzas and include a pepsi

2: five small \textbf{spinach} and bacon pizzas and a pepsi

3: please place my order for five small \textbf{spinach} and bacon pizzas with a pepsi

4: put my order in for five small \textbf{spinach} and bacon pizzas with a pepsi

\end{Verbatim}
\caption{
\claspRS: Replace Slots and Generate Text.
In this example from the \pizza dataset, we have replaced the value of Topping \underline{mushroom}
with Topping \textbf{spinach}.
The model sees $c$=5 context examples, the last of which is the original utterance,
and is prompted to generate text matching the parse with the replaced slot.
The model generates reasonable paraphrases, including the requested slots.
In particular, the model can both mix and match carrier phrase components from the
prompted examples (e.g. ``include a pepsi'') and generate novel carrier phrases,
(e.g. ``please place my order'')
presumably relying on general language knowledge acquired during
unsupervised pre-training. Note that ``[CLM]'' is a special token which the model
expects during in-context learning.
}
\label{fig:example_pizza_en2}
\end{figure}

\newpage

\subsection{Example of \claspTS: Translate Slots and Generate Text}
\label{sec:mtop_translate_slot_example}

We show an example of \claspTS (Translate Slots and Generate Text) in Figure \ref{fig:clasp_ts_mtop}.

\begin{figure*}[h!]
\begin{Verbatim}[fontsize=\scriptsize, frame=single, commandchars=\\\{\}]
\textbf{INPUT:}
[CLM] \textcolor{red}{Semantic Parse:} [IN:CREATE_REMINDER 
    [SL:PERSON_REMINDED me ]
    [SL:TODO [IN:GET_TODO [SL:DATE_TIME 10 : 00 am ] [SL:TODO doctor 's appointment ] ] ] ]; 
\textcolor{teal}{Translation in English:}
  Remind me of my 10 : 00 am doctor 's appointment; 
\textcolor{red}{Semantic Parse:} [IN:CREATE_REMINDER
    [SL:PERSON_REMINDED moi ] 
    [SL:TODO [IN:GET_TODO [SL:DATE_TIME de 10 h ] [SL:TODO rendez - vous chez le médecin ] ] ] ]; 
\textcolor{violet}{Translation in French:}
  Fais - moi penser à mon rendez - vous de 10 h chez le médecin;
\textcolor{red}{Semantic Parse:} [IN:SEND_MESSAGE 
    [SL:RECIPIENT [IN:GET_CONTACT [SL:CONTACT_RELATED \emph{my} ] [SL:TYPE_RELATION \textbf{husband} ] ] ]
    [SL:CONTENT_EXACT \underline{pick up bread} ]; 
\textcolor{teal}{Translation in English:}
  Send a message to my husband reminding him to pick up bread; 
\textcolor{red}{Semantic Parse:} [IN:SEND_MESSAGE 
    [SL:RECIPIENT [IN:GET_CONTACT [SL:CONTACT_RELATED \emph{mon} ] [SL:TYPE_RELATION \textbf{mari} ] ] ] 
    [SL:CONTENT_EXACT \underline{prendre du pain} ] ];
\textcolor{violet}{Translation in French:}

\textbf{OUTPUTs:}

0: Envoie un message à \emph{mon} \textbf{mari} pour lui rappeler de \underline{prendre du pain}

1: Envoie un message à \emph{mon} \textbf{mari} lui demandant de \underline{prendre du pain}

2: Envoie un message à \emph{mon} \textbf{mari} lui rappelant de \underline{prendre du pain}
\end{Verbatim}
% \vspace*{-0.4cm}
	\caption{
	\claspTS (Translate Slots and Generate Text).
	We first translate the slots replacing them in the parse,
	then use a one-shot in-context prompt to instruct the 20B model
	to generate a matching sentence in the target language.
	}
\label{fig:clasp_ts_mtop}
\end{figure*}

\newpage

\subsection{Example of \claspGB: Generate Both Parse and Text}

\label{sec:example_pizza_gb}
An example is shown in Figure \ref{fig:clasp_gb}

\begin{figure}[h!]
\begin{Verbatim}[fontsize=\scriptsize, frame=single, commandchars=\\\{\}]
\textbf{INPUT:}
[CLM] \textcolor{red}{Semantic Parse:} (Order
  (Pizzaorder (Number a ) (Size small ) (Topping peppers ) (Topping sausage ) (Topping pineapple ) ) )
=> \textcolor{teal}{Translation in English:} 
  can you get me a small pizza with peppers and sausage and pineapple please;
\textcolor{red}{Semantic Parse:} (Order 
  (Pizzaorder (Number a ) (Size large ) (Topping olives ) 
    (Complex_topping (Quantity extra ) (Topping cheese ) ) (Topping chicken ) ) )
=> \textcolor{teal}{Translation in English:} 
  i need a large pizza and i want olives and extra cheese as well as chicken on it thanks a lot;
\textcolor{red}{Semantic Parse:} (Order 
  (Pizzaorder (Number a ) (Size medium ) (Topping onions ) (Topping tuna ) (Topping ham ) ) )
=> \textcolor{teal}{Translation in English:}
  i'd like a medium pizza with onions tuna and ham;
\textcolor{red}{Semantic Parse:} (Order 
  (Pizzaorder (Number two ) (Topping olive ) (Topping pineapple ) (Topping mushroom ) ) )
=> \textcolor{teal}{Translation in English:}
  i want two olive pineapple and mushroom pies;
\textcolor{red}{Semantic Parse:} (Order 
  (Pizzaorder (Number a ) (Size large ) (Topping ham ) (Topping peppers ) (Not (Style thin crust ) ) ) )
=> \textcolor{teal}{Translation in English:}
  good evening how are you do me a favor and get me a large pizza with ham and peppers 
    i definitely do not want thin crust thanks;
\textcolor{red}{Semantic Parse:}

\textbf{OUTPUTS:}

0: (Order (Pizzaorder (Number two ) (Topping olive ) (Topping pineapple ) (Topping mushroom ) 
     (Not (Style thin crust ) ) ) => 
   \textcolor{teal}{Translation in English:} 
     can you get me two olive pineapple and mushroom pies please no thin crust;

1: (Order (Pizzaorder (Number a ) (Size medium ) (Topping tuna ) (Topping chicken ) ) ) => 
   \textcolor{teal}{Translation in English:}
     hello how are you i want a medium pizza with tuna and chicken on it thanks;

2: (Order (Pizzaorder (Number a ) (Size small ) (Topping chicken ) (Topping bacon ) ) ) => 
   \textcolor{teal}{Translation in English:} 
     can you please bring me a small pizza with chicken and bacon on it thanks;

3: (Order (Pizzaorder (Number a ) (Size large ) (Topping mushroom ) 
     (Topping pepperoni ) (Topping green pepper ) ) ) => 
   \textcolor{teal}{Translation in English:} 
     how are you today i want a large pizza with mushrooms pepperoni green peppers and \underline{cheese} thanks;
\end{Verbatim}
% \vspace*{-0.4cm}
	\caption{
	\claspGB{}: AlexaTM 20B generates both the parse and matching text
	remarkably well given that it has only seen five examples
	in the prompt. Intuitively this is this most ambiguous of the tasks,
	as the model has freedom to generate both a new parse and a new text.
	Therefore, we see a bit noisier generation in this mode, as shown in the example
	where an extra slot value ``cheese'' appears in the text but not in the parse.
	We filter out such examples with a heuristic string match check.
	}
\label{fig:clasp_gb}
\end{figure}

% \newpage

\subsection{Example of In-Context Sentence Translation}
\label{sec:example_in_context_mt_sent}

\begin{figure*}[h!]
    \centering
    \includegraphics[width=0.7\textwidth]{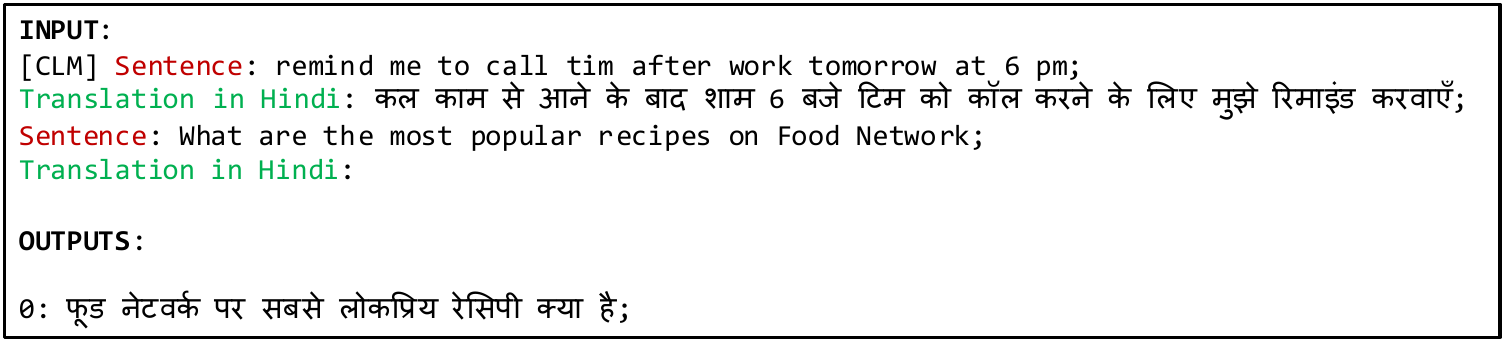}
    \caption{An example of in-context Sentence text translation from English to Hindi.}
    \label{fig:example_in_context_text_mt}
\end{figure*}

An example is shown in Figure \ref{fig:example_in_context_text_mt}.

% \newpage
\section{Data Preprocessing}

We discuss preprocessing for each of our datasets.

\subsection{Data Preprocessing for Pizza}
\label{sec:data_pre_pizza}

We provide more details about our modified Canonical Form (CF) training data, as introduced in Section \ref{sec:pizza_dataset}.
We compare the Canonical Forms released by \citet{rongali2022training} \footnote{https://github.com/amazon-research/resource-constrained-naturalized-semantic-parsing}
with the original pizza text, TOP, and EXR released by \citet{pizzaDataset}.\footnote{https://github.com/amazon-research/pizza-semantic-parsing-dataset}
An example is shown in Figure \ref{fig:example_fix_cf}, where we see that in the original data release,
EXR does not preserve the sibling order of nodes in the tree.
It appears that the CF of \citet{rongali2022training} follows the EXR, so it inherits this mismatch.

We hypothesize that this mismatch in sibling order creates an extra challenge for the model
to learn at training time, and limits the power of the naturalization approach
proposed by \citet{rongali2022training}.
In particular, in the 16-shot setting, we find that
12 out of the 16 utterances have a canonical form that does not match the original sibling order.

Observing that the sibling order is still faithfully represented in the ``TOP''
field of the corresponding utterances in the Pizza dataset \citep{pizzaDataset},
we re-produce the CF from TOP directly, using the same codebase as
\citet{rongali2022training}.\footnote{We thank the authors of \citet{pizzaDataset} and \citet{rongali2022training}
for providing support on the \pizza dataset.}
Note, we only perform this change during training time.
At testing time, we follow \citep{rongali2022training} and use the standard
grammar to parse the model output and compare using Unordered Exact Match (UEM) against 
the ground-truth EXR
(entity resolved) format.

As shown in Section \ref{sec:pizza_results}, our fixed Canonical Form
provides a very large improvement across all runs, in particular increasing
UEM from 82.54/21.00 to 90.05/58.00 on $n$=348/$n$=16, respectively.
This represents 7.51/37.00 points absolute improvement, respectively.

\begin{figure}[h!]
\begin{Verbatim}[fontsize=\scriptsize, frame=single, commandchars=\\\{\}]
Text in Arkoudas et al. (2021):
  can you get me a pizza with peppers and \textbf{sausage} and \underline{pineapple} please

TOP in Arkoudas et al. (2021):
  (ORDER can you get me (PIZZAORDER (NUMBER a ) (SIZE small ) pizza with 
      (TOPPING peppers ) and (TOPPING \textbf{sausage} ) and (TOPPING \underline{pineapple} ) ) please )

TOP-Decoupled we produced using code at Arkoudas et al. (2021):
  (ORDER (PIZZAORDER (NUMBER a ) (SIZE small ) 
      (TOPPING peppers ) (TOPPING \textbf{sausage} ) (TOPPING \underline{pineapple} ) ) )

EXR in Arkoudas et al. (2021):
  (ORDER (PIZZAORDER (NUMBER 1 ) (SIZE SMALL ) 
      (TOPPING PEPPERS ) (TOPPING \textcolor{red}{\underline{PINEAPPLE}} ) (TOPPING \textcolor{red}{\textbf{SAUSAGE}} ) ) )

Rongali et al. (2022) CF for this utterance:
  i want one small pizza with peppers , \textcolor{red}{\underline{pineapple}} , and \textcolor{red}{\textbf{sausage}}

Our CF: 
  i want one small pizza with peppers , \textcolor{teal}{\textbf{sausage}} , and \textcolor{teal}{\underline{pineapple}}
\end{Verbatim}
\caption{
Comparing our ``Fixed'' Canonical Form (``Our CF'') to the original provided by \citet{rongali2022training}.
We use the same code to resolve, we just start with the TOP and TOP-Decoupled versions provided in the dataset,
which maintain the ordering of slots in the original.
}
\label{fig:example_fix_cf}
\end{figure}

\newpage
\subsection{Data Preprocessing for mTOP}
\label{sec:data_pre_mtop}

We describe two data pre-processing steps for mTOP: (1) Space-joined Tokens, and (2) Sentinel Words.
As shown in Table \ref{tab:mtop_tokens_results}, these steps have minimal impact on non-English languages when
training on ALL data (from 73.4 to 73.5), however
\textbf{improve lower bound cross-lingual zero-shot by 17.0 points} (from 28.3 to 45.3).
Furthermore, our data pre-processing provides a moderate improvement on English, of
0.8 points (from 82.3 to 83.1) when training on en-only data,
and 0.9 points (from 82.4 to 83.3) when training on ALL data.

\begin{table}[h!]
\centering
\footnotesize
\begin{tabular}{cc | c| c | cccc| c}
\hline
 Data    & Input Source        & Word Sentinels   &   en &   de &   es &   fr &   hi &   avg-0s \\
\hline
 \multirow{3}{*}{en-only} & Utterance           & no               & 82.3 & 31.8 & 28.5 & 32.7 & 20.3 &     28.3 \\
 & Space-joined Tokens & no               & 82.9 & 34.6 & 36.6 & 39.8 & 22.8 &     33.4 \\
 & \textbf{Space-joined Tokens} & \textbf{yes}              & \textbf{83.1} & 47.3 & 51.0 & 54.8 & 28.2 &     \textbf{45.3} \\
\hline
 \multirow{3}{*}{ALL}     & Utterance           & no               & 82.4 & 71.3 & 77.4 & 74.9 & 70.1 &     73.4 \\
 & Space-joined Tokens & no               & 82.0 & 71.7 & 76.7 & 75.1 & 68.7 &     73.0 \\
 & \textbf{Space-joined Tokens} & \textbf{yes}              & \textbf{83.3} & 70.3 & 77.3 & 75.9 & 70.5 &     \textbf{73.5} \\
\hline
\end{tabular}
\caption{Results for cross-lingual zero-shot and ALL languages training on mTOP, comparing using
Utterance or space-joined tokens as input text.
In each case, the same format is used at both train and test time.}
\label{tab:mtop_tokens_results}
\end{table}

\subsubsection{Space-joined Tokens for mTOP}
\label{sec:sjt_mtop}

As noted in section \ref{sec:pizza_dataset}, the mTOP dataset\footnote{\url{https://fb.me/mtop_dataset}}
provides two options for the input: raw ``Utterance'', as well as
``tokens'', which according to the README file: 
``This is a JSON string representing the tokenization used for all experiments in the paper.''
We opt for using the provided tokens JSON, and joining the tokens on spaces.
This fixes many (although not all) spacing and other anomolies with exact match and token copying
which occur in as much as 30\% of utterances the non-English datasets.
An example for French is shown in Figure \ref{fig:example_mtop_fr}.

We encourage the community to continue a deep dive into anomalies in the mTOP dataset,
and develop a standard setting, perhaps even releasing a standardized / cleaned mTOP-v2.
As it stands, we still consider mTOP a highgly useful dataset to evaluate experiments within the same publication or research team, however comparisons across publications and groups should be taken with a grain of salt.

\begin{figure}[h!]
\begin{Verbatim}[fontsize=\scriptsize, frame=single, commandchars=\\\{\}]
Utterance field in mTOP French:
  Donne-moi la liste des salons de \textcolor{red}{\textbf{l'automobile}} prévus à Atlanta le \textcolor{red}{\underline{week-end}} prochain

Ground-truth parse:
  [IN:GET_EVENT [SL:CATEGORY_EVENT salons de \textcolor{teal}{\textbf{l' automobile}} ] [SL:LOCATION Atlanta ]
    [SL:DATE_TIME le \textcolor{teal}{\underline{week - end}} prochain ] ]

\textbf{Space-joined tokens field (our models use this version)}:
  Donne - moi la liste des salons de \textcolor{teal}{\textbf{l' automobile}} prévus à Atlanta le \textcolor{teal}{\underline{week - end}} prochain

\end{Verbatim}
\caption{
Comparing ``space-joined tokens'' input versus ``Utterance'' input format for mTOP.
As shown, the ``space-joined-tokens'' resolves various spacing anomalies 
which improves cross-lingual zero-shot performance.
}
\label{fig:example_mtop_fr}
\end{figure}

% provides both the original text, and the tokens used by the
% pre-tokenization step when preparing the data.
% For example, time expressions are often space-separated, so whereas the original text might have
% "2:00", the parse will have "2 : 00".
% We found this to cause up to 30\% of the data to have a mismatch in the formatting
% between the text and the parse.

% Therefore, we opted to construct the text input by space-joining the provided tokens
% in the dataset.
% We applied the same procedure for the train, eval, and test splits.
% We found this to reduce the occurrence of mismatch to less than 3\%,
% and as shown in the Appendix (TODO),
% it improved zero-shot performance significantly.

\subsubsection{Sentinel Words for mTOP}
\label{sec:sentinel_words_mtop}

Following \citet{raman-etal-22-st-s2s}, we use ``sentinel words'' which we show greatly improves the cross-lingual
zero-shot performance. An example is shown in Figure \ref{fig:sentinel_words_mtop}.

As noted in section \ref{sec:data_pre_mtop}, we use \textit{Space-joined Tokens} as input,
which resolves many spacing anomalies occurring in the ground-truth annotation for a
large portion (up to 30\% of non-English) of the data.
Still, approximately 3\% of the non-English data has unresolved spacing and casing anomalies
(see also, Appendix \ref{sec:sciem_metric}).
In those cases, we simply discard the original training utterances which cannot be converted into sentinel form.
When an unresolved spacing or casing anomaly occurs in a test utterance,
we do not discard the the utterance, but rather use a metric
which makes it possible for the model to 
recover the correct answer (see Appendix \ref{sec:sciem_metric}).

We do not add these sentinel words to the vocabulary,
but rather simply allow the sentencepiece \citep{kudo-richardson-2018-sentencepiece} tokenizer to split
them into subwords, such as \texttt{['\_word', '0']}.
We hypothesize that this could allow the model to generalize at inference time to inputs
longer than those seen during training.
However, this choice makes the input and output sequences longer than necessary,
which could impact latency.
In future work, we would like to explore adding the sentinel words to the
vocabulary and measure this trade-off explicitly.

\begin{figure}[h!]
\begin{Verbatim}[fontsize=\scriptsize, frame=single, commandchars=\\\{\}]
## English example ##
Original Text:
  are there \textbf{thunder storms} on the forecast \underline{this} \underline{weekend}
  
Original Parse:
  [IN:GET_WEATHER [SL:WEATHER_ATTRIBUTE \textbf{thunder storms} ] [SL:DATE_TIME \underline{this} \underline{weekend} ] ]
----------
Sentinel Words Text:
  word0 are word1 there \textbf{word2} thunder \textbf{word3} storms word4 on word5 the
    word6 forecast \underline{word7} this \underline{word8} weekend

Sentinel Words Parse:
  [IN:GET_WEATHER [SL:WEATHER_ATTRIBUTE \textbf{word2} \textbf{word3} ] [SL:DATE_TIME \underline{word7} \underline{word8} ] ]
----------

## German example ##
Original Text:
  Sind \underline{für} \underline{dieses} \underline{Wochenende} \textbf{Gewitter} vorhergesagt ?

Original Parse:
  [IN:GET_WEATHER [SL:WEATHER_ATTRIBUTE \textbf{Gewitter} ] [SL:DATE_TIME \underline{für} \underline{dieses} \underline{Wochenende} ] ]
----------
Sentinel Words Text:
  word0 Sind \underline{word1} für \underline{word2} dieses \underline{word3} Wochenende \textbf{word4} Gewitter word5 vorhergesagt word6 ?

Sentinel Words Parse:
  [IN:GET_WEATHER [SL:WEATHER_ATTRIBUTE \textbf{word4} ] [SL:DATE_TIME \underline{word1} \underline{word2} \underline{word3} ] ]
----------
\end{Verbatim}
\caption{
An example of the input and output formats when using sentinel words.
}
\label{fig:sentinel_words_mtop}
\end{figure}

% \begin{table}[h!]
% \footnotesize
% \centering
% \begin{tabular}{cc | c| cccc| c}
% \hline
%  Data    & Sentinel Words   &   en &   de &   es &   fr &   hi &   avg-0s \\
% \hline
%  \multirow{2}{*}{en-only} & no               & 82.8 & 34.6 & 35.5 & 38.6 & 23.0 &     32.9 \\
%  & yes              & 83.1 & 46.9 & 50.0 & 53.1 & 27.9 &     \textbf{44.5} \\
% \hline
%  \multirow{2}{*}{ALL}     & no               & 82.2 & 70.4 & 75.9 & 72.9 & 71.5 &     \textbf{72.7} \\
%       & yes              & 83.1 & 69.7 & 75.7 & 73.8 & 70.5 &     72.4 \\
% \hline
% \end{tabular}
% \caption{The impact of Sentinel Words on mTOP SCIEM Exact Match.}
% \label{tab:sentinel_words_results}
% \end{table}

\section{Space- and Case-Insensitive Exact Match (SCIEM) Metric for mTOP}
\label{sec:sciem_metric}

We define the variant of Exact Match we use for mTOP,
which we call Space- and Case-Insensitive Exact Match (SCIEM).
SCIEM is \textit{insensitive to spacing and casing of text words in the parse}
(excluding the parse elements such as the intent and slot names).
Python code is provided in Figure \ref{fig:sciem_code} and an example is shown in Figure \ref{fig:sciem_example}.
\textbf{We encourage the research community to adopt these standard settings for mTOP:
Space-joined Tokens as Input, and SCIEM metric.}

We compare results using Verbatim Exact Match vs. SCIEM, with greedy decoding (``Greedy''),
in Table \ref{tab:sciem_results}.
As show in the table, SCIEM provides a small boost in performance on the non-English languages,
of 0.5 points on 
the lower bound ``en-only'' (from 44.5 to 45.0),
0.9 points on the upper bound ``ALL'' (from 72.4 to 73.3),
0.7 points on our baseline method ``MT-Opus'' (from 59.5 to 60.2),
and 0.8 points on our best-performing combination of methods ``Our Best'' (from 65.4 to 66.2).

Note, however, that the difference is unequal across languages, e.g. in the ``en-only'' setting,
switching from Verbatim Exact Match to SCIEM improves French (``fr'') by 1.1 points (from 53.1 to 54.2)
however does not impact Hindi (``hi'') at all.
Finally, SCIEM has minimal impact on ``en'' results, with ``ALL'' improving by 0.2 points (from 83.1 to 83.3)
and the other settings matching exactly.

These trends match with our observations in Appendices \ref{sec:sjt_mtop} and \ref{sec:sentinel_words_mtop},
that even after using space-joined tokens and sentinel words for the input, there remain a small number
of spacing and casing anomalies, some of which are resolved by using the SCIEM metric.

\begin{figure}[h!]
\begin{Verbatim}[fontsize=\scriptsize, frame=single, commandchars=\\\{\}]
def get_sciem_key(model_output):
    pieces = model_output.strip().split()
    new_pieces = []
    for piece in pieces:
        if piece.startswith('[IN:') or piece.startswith('[SL:') or piece == ']':
            new_pieces.append(piece)
        else:
            new_pieces.append(piece.lower())
    return ''.join(new_pieces)

>>> model_output = "[IN:GET_WEATHER [SL:DATE_TIME para el Domingo de Pascua a las 14 : 00] ]"
>>> get_sciem_key(model_output)
'[IN:GET_WEATHER[SL:DATE_TIMEparaeldomingodepascuaalas14:00]]'
\end{Verbatim}
\caption{Python code for SCIEM metric.}
\label{fig:sciem_code}
\end{figure}

\begin{figure}[h!]
\begin{Verbatim}[fontsize=\scriptsize, frame=single, commandchars=\\\{\}]
Example from mTOP Spanish
----------
\textit{Utterance} Input:
  Dime el pronóstico para el \textcolor{red}{\textbf{Domingo}} de Pascua a las \underline{14:00}.

\textit{Space-joined Tokens} Input:
  Di me el pronóstico para el \textcolor{red}{\textbf{Domingo}} de Pascua a las \underline{14 : 00} .
----------
Model hypothesis when using \textit{Utterance}:
  [IN:GET_WEATHER [SL:DATE_TIME para el \textcolor{red}{\textbf{Domingo}} de Pascua a las \underline{14:00} ] ]

Model hypothesis when using \textit{Space-joined Tokens}:
  [IN:GET_WEATHER [SL:DATE_TIME para el \textcolor{red}{\textbf{Domingo}} de Pascua a las \underline{14 : 00} ] ]

Ground-truth Parse Original:
  [IN:GET_WEATHER [SL:DATE_TIME para el \textcolor{teal}{\textbf{domingo}} de Pascua a las \underline{14 : 00} ] ]
----------
Model Hypothesis (in both cases) For Space- and Case-Insensitive Exact Match (SCIEM):
  [IN:GET_WEATHER[SL:DATE_TIMEparael\textcolor{teal}{\textbf{domingo}}depascuaalas\underline{14:00}]]

Ground-truth Parse For Space- and Case-Insensitive Exact Match (SCIEM):
  [IN:GET_WEATHER[SL:DATE_TIMEparael\textcolor{teal}{\textbf{domingo}}depascuaalas\underline{14:00}]]
----------
Verbatim Exact Match? NO
SCIEM Exact Match? YES
\end{Verbatim}
\caption{
An example of Space- and Case-Insensitive Exact Match (SCIEM).
The original \textit{Utterance} input has both a spacing (``14:00'' vs. ``14 : 00'') and a casing (``Domingo'' vs. ``domingo'') anomaly
compared to the Ground-truth Parse.
While using \textit{Space-joined Tokens} as input solves the spacing issue, the casing issue remains.
In both cases, SCIEM corrects for the anomalies in the test set by counting the model's hypothesis as correct.
}
\label{fig:sciem_example}
\end{figure}

\begin{table}[h!]
\centering
\footnotesize
\begin{tabular}{ccc | c | cccc | c}
\hline
 Data    & Decoding   & Exact Match Type   &   en &   de &   es &   fr &   hi &   avg-0s \\
\hline
  \multirow{3}{*}{en-only} & Greedy     & Verbatim          & 83.1 & 46.9 & 50.0 & 53.1 & 27.9 &     44.5 \\
       & Greedy     & SCIEM             & 83.1 & 47.2 & 50.8 & 54.2 & 27.9 &     45.0 \\
       & \textbf{Beam4}      & \textbf{SCIEM}             & 83.1 & 47.3 & 51.0 & 54.8 & 28.2 &     \textbf{45.3} \\
\hline 
  \multirow{3}{*}{ALL}     & Greedy     & Verbatim          & 83.1 & 69.7 & 75.7 & 73.8 & 70.5 &     72.4 \\
       & Greedy     & SCIEM             & 83.3 & 70.2 & 77.0 & 75.6 & 70.5 &     73.3 \\
       & \textbf{Beam4}      & \textbf{SCIEM}             & 83.3 & 70.3 & 77.3 & 75.8 & 70.5 &     \textbf{73.5} \\

 \hline
 \multirow{3}{*}{MT-Opus} & Greedy     & Verbatim          & 82.9 & 63.2 & 63.9 & 63.5 & 47.3 &     59.5 \\
       & Greedy     & SCIEM             & 82.9 & 63.5 & 64.9 & 64.9 & 47.3 &     60.2 \\
       & \textbf{Beam4}      & \textbf{SCIEM}             & 83.0 & 63.8 & 65.0 & 65.1 & 47.4 &     \textbf{60.3} \\
 \hline
 \multirow{3}{*}{\makecell[l]{(Our Best) \claspTSTB\\~~~~~~~+MT-20B +MT-Opus}}    & Greedy     & Verbatim          & 84.4 & 66.1 & 66.7 & 70.8 & 57.9 &     65.4 \\
       & Greedy     & SCIEM             & 84.4 & 66.5 & 67.9 & 72.4 & 57.9 &     66.2 \\
       & \textbf{Beam4}      & \textbf{SCIEM}             & 84.4 & 66.7 & 68.1 & 72.6 & 58.1 &     \textbf{66.4} \\
\hline
\end{tabular}
\caption{The impact of SCIEM (vs. Verbatim Exact Match) and Beam4 decoding (vs. Greedy decoding)
on lower bound (``en-only''), upper bound (``ALL''), baseline (``MT-Opus''), and our best-performing (``Our Best'') 
combination of methods.}
\label{tab:sciem_results}
\end{table}

\section{Impact of Test-Time Decoding Strategy}
\label{sec:ablation_beam}

In Table \ref{tab:sciem_results} (Appendix \ref{sec:sciem_metric}), we also compare the impact of our
choice of Decoding Strategy. As show in the Table,
across settings Beam4 provides only a small boost over Greedy decoding, between 0.1 and 0.3 points on ``avg-0s'',
and either exactly the same or 0.1 points improvement on ``en''.

\section{Impact of Adding Grammar-Generated Train Data for \pizza}
\label{sec:learning_curves}

\begin{figure}[h!]
\centering
\begin{subfigure}{.49\textwidth}
  \centering
  \includegraphics[width=0.9\textwidth]{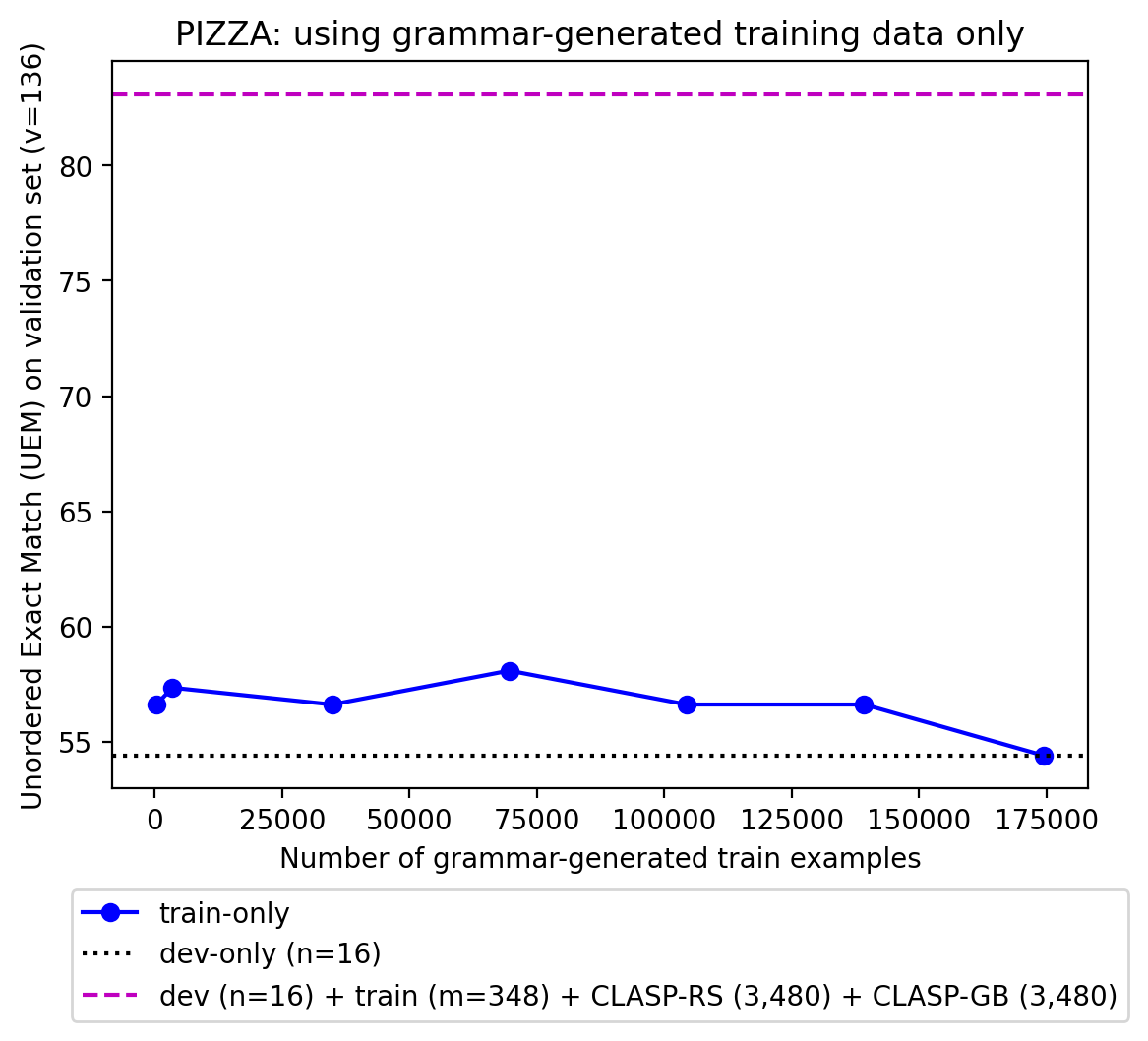}
\caption{
Training data in isolation.
}
  \label{fig:sub1}
\end{subfigure}%
% \hspace{0.05 \textwidth}
\hfill
\begin{subfigure}{.49\textwidth}
  \centering
\includegraphics[width=0.9\textwidth]{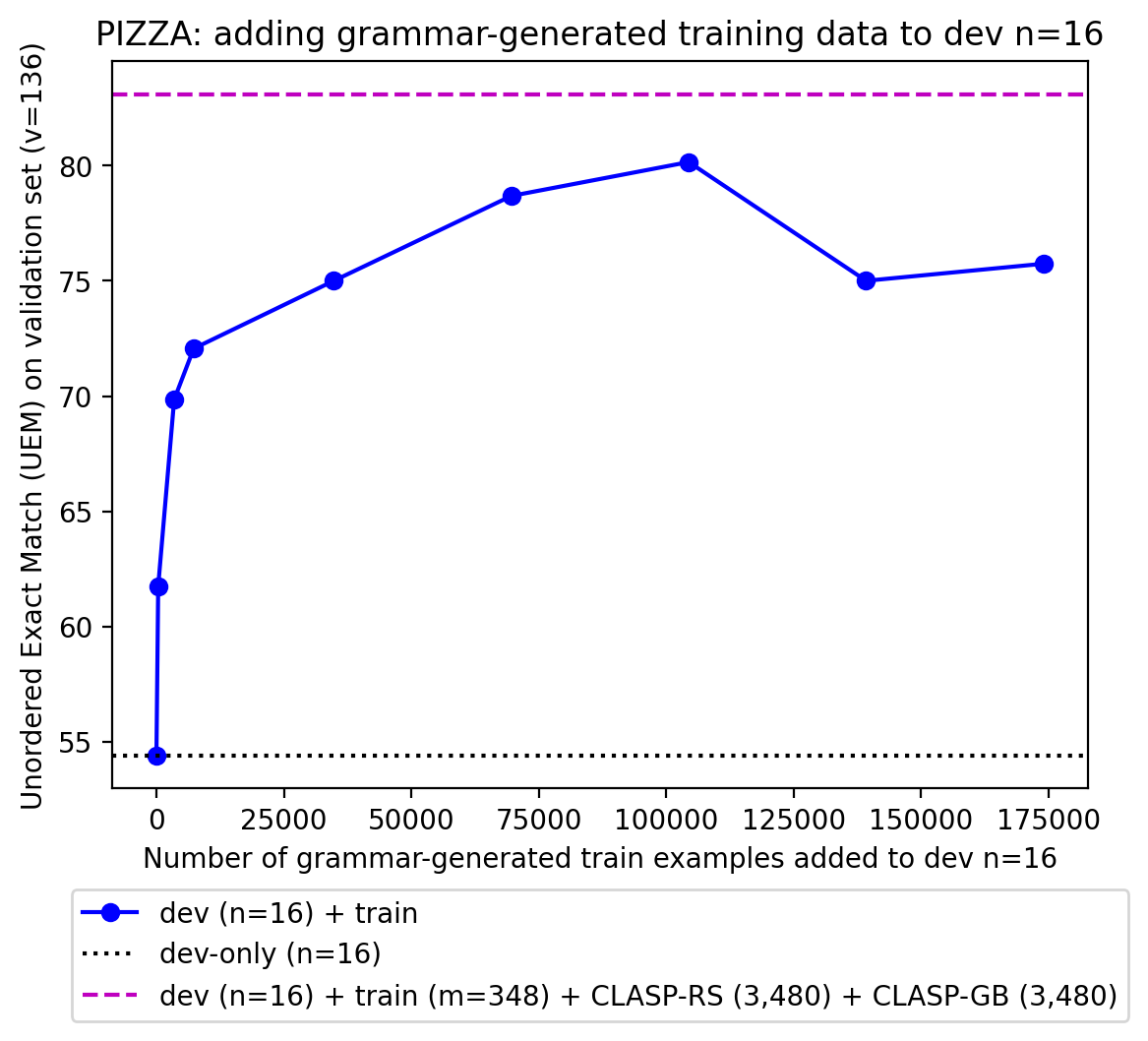}
\caption{
Training data mixed with human dev (n=16) data.
}
  \label{fig:sub2}
\end{subfigure}
\caption{
Learning Curve of increasing amount of (grammar-generated) training data
for \pizza{}. Left (a) in isolation; Right (b) mixed with (human-curated) dev n=16.
}
\label{fig:lc_pizza_n16}
\end{figure}

For \pizza, we show the impact on tuning the amount of grammar-generated training data,
as described in Section \ref{sec:baselines}.
As show in Figure \ref{fig:lc_pizza_n16}, the best-performing option for train (m) in isolation is m=69,600, and when mixed with
dev (n=16) + train (m), m=104,400 is best.
These correspond to the rows ``train-only'' and ``dev+train'', respectively, in table \ref{tab:pizza_results}.
Note, as described in Section \ref{sec:pizza_dataset}, to avoid overfitting
on the test set which contains only 1,357 utterances,
we extract a 10\% subset of the test set, referred to as the ``validation''
set to use for hyperparameter tuning and early stopping.

% Subfigure!!!

\section{Hyperparameters}
\label{sec:hyperparams}

We fine-tune with Adam \citep{Kingma2015AdamAM} using a learning rate $1e-5$,
dropout 0.1, and batch size 128. We fix the number of update steps to
$u$=2,500 (1,000 epochs for dev $n$=348 or 20,000 epochs for dev $n$=16) for \pizza, and $u$=12,000 (100 epochs) for mTOP.
Fine-tuning takes takes one hour for \pizza and four hours for mTOP on
an AWS p3.24xlarge instance, using DeepSpeed ZeRO \cite{deepspeed} Stage 1 to save GPU memory
and speed up training. Our models are built on top of HuggingFace \cite{wolf-etal-2020-transformers}.

When generating data with AlexaTM 20B, we use either sampling or greedy decoding, described in Appendix \ref{sec:filtering}.

\section{mTOP Utterances Used for Prompting}

\label{sec:prompt_utts}

\begin{figure*}[h!]
    \centering
    \includegraphics[width=0.55\textwidth]{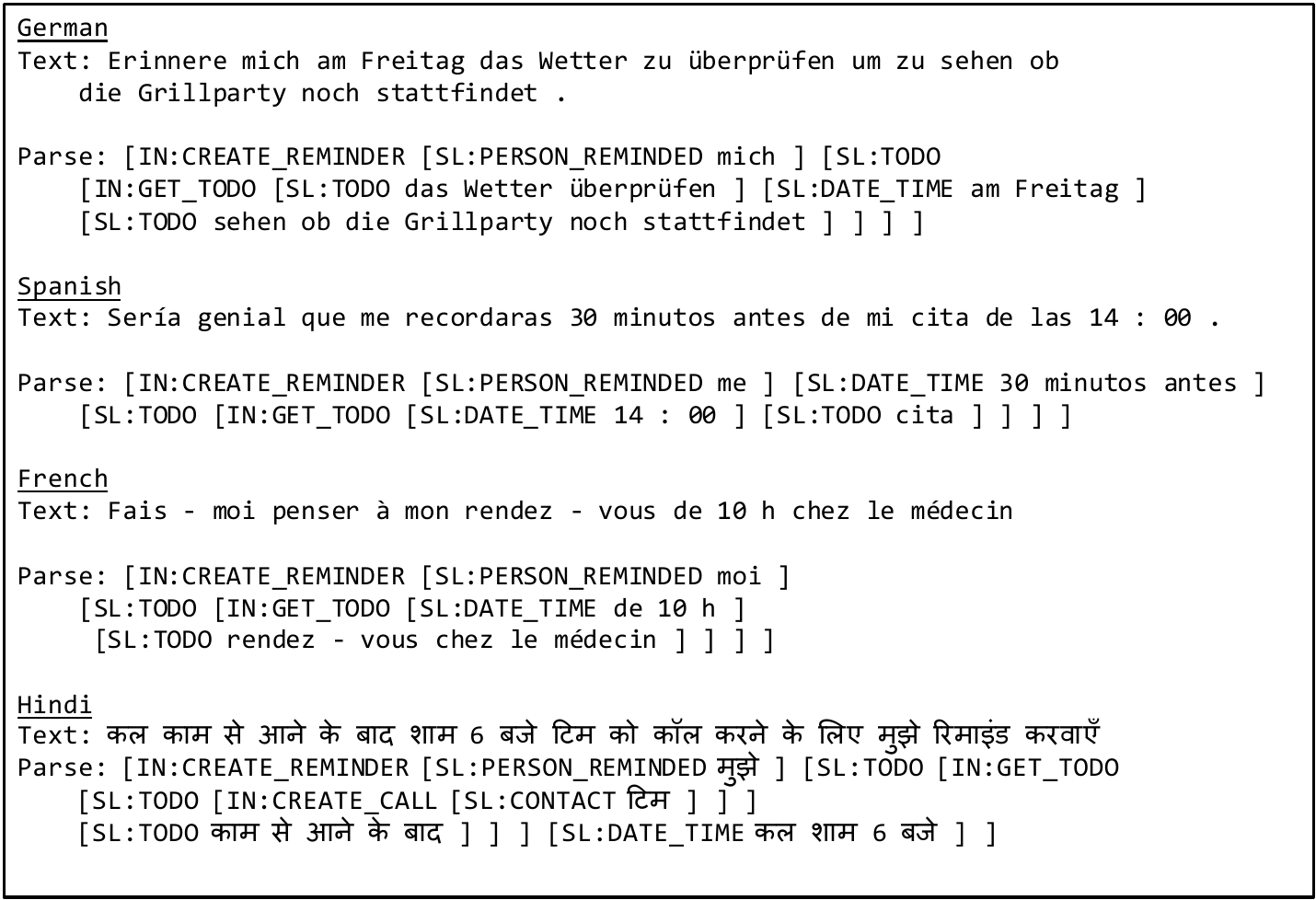}
    \caption{The one-shot examples from mTOP which we
    use for all in-context prompts.}
    \label{fig:mtop_samples}
\end{figure*}

The utterances we use for all mTOP in-context generation prompts are shown in Figure \ref{fig:mtop_samples}.

\section{Filtering \clasp Outputs}
\label{sec:filtering}

Our filtering logic starts from the following two Validation Principles:
\textit{VP1 (Valid Parse)}: the parse must be valid according to the task format and the specific instructions contained in the generation prompt (e.g. including a particular slot);
\textit{VP2 (All Slots Present)}: each slot value in the parse must appear in the sentence text.

\subsection{Filtering \clasp Outputs for \pizza}
\label{sec:filtering_pizza}

For \textbf{\pizza}, we generate 4 outputs with
sampling\footnote{We refer the reader to this guide: https://huggingface.co/blog/how-to-generate .}
(settings: $top\_k=50$ \cite{fan-etal-2018-hierarchical}, $top\_p=0.9$ \citep{Holtzman2020TopP}, and $temperature=0.9$),
discard any which are invalid according to certain heuristic Failure Modes (described below),
then select the remaining one with lowest perplexity.
In cases where there is no acceptable output utterance,
we duplicate an utterance from the prompt
back into the training set to maintain the per-class distribution.

We define the \textbf{Success Rate (Inputs) as the percentage of
input prompts which result in at least one valid output}.
In early experiments, we used the Success Rate (Inputs) metric to iterate
on settings such as the the number of input examples, the prompt format,
and the sampling hyperparameters.
Our final settings produce a Success Rate (Inputs) of
\textbf{81.1\% for \claspRS} (Replace Slots then Generate Text; Section \ref{sec:method_replace_slots}) and
\textbf{77.6\% for \claspGB} (Generate Both Parse and Text; Section \ref{sec:method_gen_both})
(Table \ref{tab:filtering_pizza}).

The lower Success Rate (Inputs) for \claspGB reflects the
greater degree of ambiguity for this \clasp method, as the model must \textit{generate both} the
the parse and text.
We also measure the Success Rate (Outputs)
as the percentage of all \textit{outputs} which are valid, and see a similar trend.

We identify a total of seven common Failure Modes, which are (non-mutually exclusive)
criteria for discarding a generated utterance.
The occurrence rate  for each is shown in Table \ref{tab:filtering_pizza},
where the denominator is the total number of \textit{outputs} produced.

The most common Failure Mode is ``Missing Slot'', where the output is missing one of the requested slot values,
occurring 25.8\%/30.0\% of the time for \claspRS/\claspGB.
``Untagged Slot'' occurs when a slot word from the catalog, such as ``pepperoni'' appears in the outputs,
but is not tagged in any slot, occurring for 1.6\%/7.1\% of outputs.
Invalid Separators (semicolon or arrow ``\texttt{=>}'' is missing from or mis-placed or duplicated in the output)
occurs for 0.1\%/2.1\% of outputs.
3.4\%/0.8\% of the outputs are discarded due to copying an input example.

We discard \textit{Duplicate Outputs}, occurring for 39.3\% of the \claspRS and 3.6\% of the \claspGB outputs,
respectively.
The higher (lower) portion of duplicates for \claspRS (\claspGB) reflects how the method is more (less) constrained,
resulting the model's ability to produce less (more) diverse outputs.

Finally, for \claspGB, we discard outputs which have an Invalid Parse or Unk. (Unknown) Entity
according to the catalog.
The \textbf{Invalid Parse percentage is remarkably low, just 0.9\%, suggesting that the
\claspGB method is effective at teaching the LLM to produce valid Semantic Parsing training data
from very few examples}.

The Unknown Entity portion of 6.3\% may represent an opportunity to expand the catalog,
either automatically or via a human annotation pipeline.
For example, in one case the model produced ``lemonade'' as a Drinktype, which
is reasonable, however was discarded since it does not appear in the slot catalogs.

Future work can discover more failure modes to filter out,
and explore methods to improve the quality of outputs so that less filtering is required.

\begin{table}[h!]
\footnotesize
\centering
\begin{tabular}{c|cc||ccccc|cc}
\hline

\multirow{3}{*}{\makecell{\clasp\\Method}} &
\multirow{3}{*}{\makecell{Success\\Rate\\(Inputs)}} & 
\multirow{3}{*}{\makecell{Success\\Rate\\(Outputs)}} &
\multicolumn{7}{c}{Failure Modes} \\\cline{4-10}
&    &    &
\makecell{Missing\\Slot}   &
\makecell{Untagged\\Slot}   &
\makecell{Invalid\\Separators}  &
\makecell{Copy\\Example} & 
\makecell{Duplicate\\Output} & 
\makecell{Invalid\\Parse} &
\makecell{Unk.\\Entity} \\

\hline
  \claspRS             &              \textbf{81.1} &              66.2 &                 25.8 &            1.6 &                  0.1 &             3.4 &        39.3 &            --   &         --   \\
 \claspGB             &              \textbf{77.6} &              34.9 &                 30.0 &            7.1 &                  2.1 &             0.8 &         3.6 &             0.9 &          6.3 \\
\hline
\end{tabular}
\caption{Success rate (percentage) and occurrence of Failure Modes (percentage)
when generating data for \pizza using the \clasp methods, \claspRS and \claspGB.
The Success rate (Inputs) for each line is bolded.}
\label{tab:filtering_pizza}
\end{table}

\subsection{Filtering \clasp Outputs for mTOP}
\label{sec:filtering_mtop}

For \textbf{mTOP}, we use greedy search which returns only one output per input prompt.
Then, similar to our setup for \pizza, we discard outputs which exhibit one or more Failure Modes (described below), and when there is no acceptable output utterance,
we duplicate an utterance from the prompt
back into the training set to maintain the per-class distribution.

We define \textbf{Success Rate as the percentage of inputs which result in a valid output}
after filtering.
As show in Table \ref{tab:filtering_mtop}, the overall Success Rate
(averaged across the four non-English languages) is
\textbf{87.9\% for \claspTS} (Translate Slots then Generate Parse, Section 
\ref{sec:method_translate_slots}) and \textbf{76.3\% for \claspTB}
(Translate both Parse and Text, Section \ref{sec:method_translate_both}).
We further analyze the Success Rate by three Success Modes: ``Clean'' (77.3\%/64.4\% for \claspTS/\claspTB)
where no post-processing is needed, and two heuristic recovery methods,
``Slot n-best'' and ``Fix Casing'', described in the next section.

Given that \claspTB is more challenging
(the model must generate not only the text but also the parse),
it is not surprising to find that the Success Rate is lower for this method
compared to the \claspTS.
However, as show in Section \ref{sec:mtop_results}, the two methods provide
similar downstream performance. This suggests
that although \claspTB provides a smaller volume of viable data than \claspTS, the data
from \claspTB is of higher quality (perhaps due to avoiding the noise of translating slots a priori).

The most common Failure Mode is ``Missing Slot'', described above for \pizza
in Appendix \ref{sec:filtering_pizza}.
While the model rarely copies an input example verbatim,
Invalid Separators (\texttt{=>} and semicolon) occur
for 12.4\% of for Hindi outputs, discussed in more detail
in Appendix \ref{sec:filtering_mt}.

Finally, while the model rarely outputs invalid parses,
we observe a high rate of the
``Mismatch Parse'' failure mode, where the output
parse does not match the input example structure.\footnote{
Note for mTOP, our goal is not to generate novel parse structures,
but rather to create a parallel dataset from English to the other languages.}
We find the majority of these occur when the model copies part of one of the input examples,
as show in Figure \ref{fig:example_mtop_failure}.
In early experiments, we found that adding more examples to the prompt exacerbated this problem,
so we decided to always use just one example.

Future work can explore how to reduce the occurrence of these failure modes
to extract even more performance boost from \clasp.

% === v2 === %
\begin{table}[h!]
\footnotesize
\centering
\begin{tabular}{c|cc|ccc||cc|ccc}
\hline
\multirow{3}{*}{\makecell{\clasp\\Method}} &
\multirow{3}{*}{\makecell{Language}} & 
\multirow{3}{*}{\makecell{Success\\Rate}} &
% \multirow{3}{*}{\makecell{Clean}} &
% \multirow{3}{*}{\makecell{Slot\\n-best}} &
% \multirow{3}{*}{\makecell{Fix\\Casing}} &
\multicolumn{3}{c||}{Success Modes} &
\multicolumn{4}{c}{Failure Modes}\\\cline{4-6}\cline{7-11}
&    &    
& 
\makecell{Clean} &
\makecell{Slot\\n-best} &
\makecell{Fix\\Casing} &
\makecell{Missing\\Slot}   &
\makecell{Copy\\Example} & 
\makecell{Invalid\\Separators}  &
\makecell{Invalid\\Parse} &
\makecell{Mismatch\\Parse} \\
%clasp Method   & Language   &   Success &   Clean &   Slot Re-translation &   Fix Casing &   Missing &   Copy Original &   Invalid Separators &   Invalid Parse &   Mismatch Parse \\
\hline
 \multirow{5}{*}{TS}             & de         &      84.4 &    72.8 &                   8.7 &          2.9 &      15.4 &             0.2 &                 --   &            --   &             --   \\
              & es         &      86.5 &    76.0 &                   6.7 &          3.7 &      13.0 &             0.5 &                 --   &            --   &             --   \\
              & fr         &      90.4 &    78.8 &                   7.2 &          4.4 &       9.4 &             0.2 &                 --   &            --   &             --   \\
              & hi         &      90.3 &    81.6 &                   8.7 &          0.0 &       9.7 &             0.0 &                 --   &            --   &             --   \\\cline{2-3}\cline{4-6}
              & avg        &      \textbf{87.9} &    77.3 &                   7.8 &          2.8 &      --   &            --   &                 --   &            --   &             --   \\
\hline
 \multirow{5}{*}{TB}             & de         &      78.8 &    70.9 &                   1.8 &          6.2 &      11.7 &             0.6 &                  0.6 &             0.9 &              7.3 \\
              & es         &      82.2 &    61.5 &                  18.3 &          2.4 &      14.3 &             1.6 &                  0.7 &             0.0 &              1.1 \\
              & fr         &      76.7 &    62.6 &                  13.0 &          1.2 &       8.0 &             1.1 &                  1.0 &             0.5 &             12.6 \\
              & hi         &      67.5 &    62.8 &                   4.6 &          0.1 &       6.0 &             0.1 &                 12.4 &             1.6 &             12.4 \\\cline{2-3}\cline{4-6}
              & avg        &      \textbf{76.3} &    64.4 &                   9.4 &          2.5 &      --   &            --   &                 --   &            --   &             --   \\
\hline
\end{tabular}
\caption{Success Rate and occurrence of various Success Modes and Failure Modes
when generating data for mTOP using the \clasp methods, \claspTS and \claspTB.
All numbers represent percentage of occurrence.
The average across the four languages for each \clasp method is bolded.}
\label{tab:filtering_mtop}
\end{table}
% === v2 === %

\begin{figure}[h!]
\begin{Verbatim}[fontsize=\scriptsize, frame=single, commandchars=\\\{\}]
\textbf{INPUT}:
[CLM] \textcolor{red}{Semantic Parse for English:}
  [IN:CREATE_REMINDER [SL:PERSON_REMINDED me ] [SL:TODO
    [IN:GET_TODO [SL:DATE_TIME 10 : 00 am ] [SL:TODO doctor 's appointment ] ] ] ]
=> \textcolor{teal}{Translation in English:}
  Remind me of my 10 : 00 am doctor 's appointment;
\textcolor{red}{Semantic Parse for French:}
  [IN:CREATE_REMINDER [SL:PERSON_REMINDED moi ] [SL:TODO
    [IN:GET_TODO [SL:DATE_TIME de 10 h ] [SL:TODO rendez - vous chez le médecin ] ] ] ]
=> \textcolor{teal}{Translation in French:}
  Fais - moi penser à mon rendez - vous de 10 h chez le médecin;
\textcolor{red}{Semantic Parse for English:}
  [IN:SET_RSVP_NO ]
=> \textcolor{teal}{Translation in English:}
  RSVP no to this event;
\textcolor{red}{Semantic Parse for French:}

\textbf{OUTPUT}:
[IN:SET_RSVP_NO 
  \underline{[SL:PERSON_REMINDED moi ] [SL:TODO [IN:GET_TODO [SL:DATE_TIME de 10 h ]}
  \underline{[SL:TODO rendez - vous chez le médecin ] ] ] ]}
=> \textcolor{teal}{Translation in French:} \underline{Fais - moi penser à mon rendez - vous de 10 h chez le médecin};

\end{Verbatim}
\caption{Example of Failure Mode Mismatch Parse for \claspTB.
While the output parse is technically valid according to the
mTOP specification, it does not match the requested parse format.
In particular, in this case, it copies part of the other example's prompt verbatim.}
\label{fig:example_mtop_failure}
\end{figure}

\subsubsection{Slot N-Best and Casing Recovery for mTOP}
\label{sec:slot_nbest}

There is inherent ambiguity of word choice in cross-lingual data generation.
When a slot has a different form in the parse vs. in the text,
the example is considered invalid (VP2, above), and would need to be discarded.
However, we identify two modes, ``Slot n-best'' and ``Fix Casing'',
where it is possible to recover from this mismatch
by simply replacing the slot value in the parse with a readily available alternative.

\begin{figure*}[h!]
    \centering
    \includegraphics[width=0.5\textwidth]{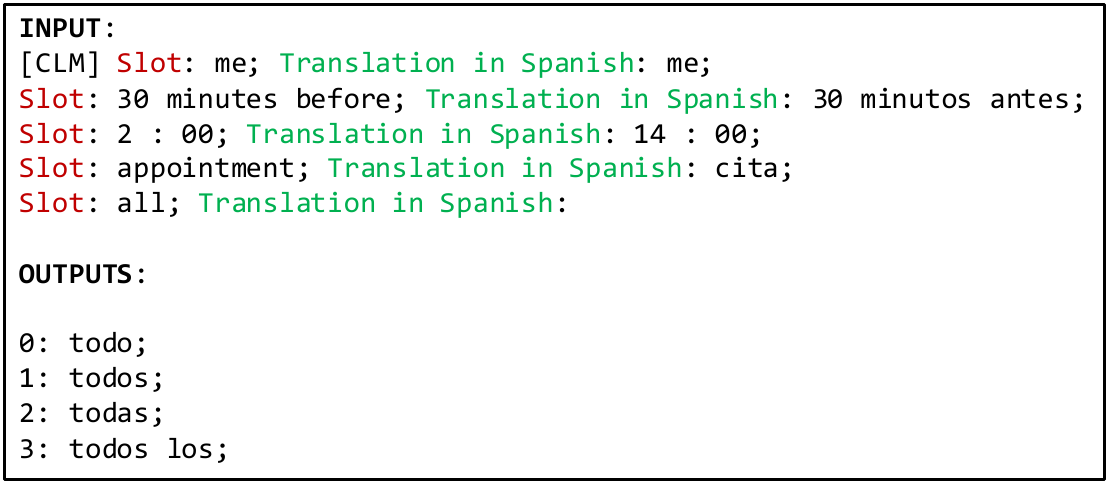}
    \caption{An example of in-context Slot text translation from English to Spanish.}
    \label{fig:example_in_context_slot_mt}
\end{figure*}

For ``\textbf{Slot n-best}'', we \textit{a priori} create an n-best list of 
all slot translations, using an in-context prompt with AlexaTM 20B (see Figure \ref{fig:example_in_context_slot_mt})
and beam search 4 outputs.
Then, as show in Figure \ref{fig:example_mtop_slot_nbest},
if we find that a slot is missing from the text,
we check for the presence of another version of the slot
from the n-best list, and if found, update
the parse with the new value, and accept the generated training example.
As show in Table \ref{tab:filtering_mtop},
this allows us to recover 7.8\%/9.4\% of Success Rate
for \claspTS/\claspTB.

Similarly, for ``\textbf{Fix Casing}'' (see Figure \ref{fig:example_mtop_fix_casing})
if we find that a slot is missing from the text,
we check for a case-insensitive match in the text,
and if found, replace the slot in the parse.
This allows us to recover 2.8\%/2.5\% of
Success Rate for \claspTS/\claspTB (Table \ref{tab:filtering_mtop}).

\begin{figure}[h!]
\begin{Verbatim}[fontsize=\scriptsize, frame=single, commandchars=\\\{\}]
\textbf{INPUT}:
[CLM] \textcolor{red}{Semantic Parse:}
  [IN:CREATE_REMINDER [SL:PERSON_REMINDED me ] [SL:DATE_TIME 30 minutes before ] 
    [SL:TODO [IN:GET_TODO [SL:DATE_TIME 2 : 00 ] [SL:TODO appointment ] ] ] ];
\textcolor{teal}{Translation in English:}
  It would be great if you could remind me 30 minutes before my 2 : 00 appointment .;
\textcolor{red}{Semantic Parse:}
  [IN:CREATE_REMINDER [SL:PERSON_REMINDED me ] [SL:DATE_TIME 30 minutos antes ]
    [SL:TODO [IN:GET_TODO [SL:DATE_TIME 14 : 00 ] [SL:TODO cita ] ] ] ];
\textcolor{teal}{Translation in Spanish:}
  Sería genial que me recordaras 30 minutos antes de mi cita de las 14 : 00 .;
\textcolor{red}{Semantic Parse:}
  [IN:GET_ALARM [SL:AMOUNT \textbf{all} ] [SL:DATE_TIME for Friday ] ];
\textcolor{teal}{Translation in English:}
  I want to see all alarms for Friday;
\textcolor{red}{Semantic Parse:}
  [IN:GET_ALARM [SL:AMOUNT \underline{todo} ] [SL:DATE_TIME viernes ] ];
\textcolor{teal}{Translation in Spanish:}

\textbf{OUTPUT}:
Quiero ver \underline{todas} las alarmas para el viernes.;

\textbf{SLOT N-BEST RECOVERY}:
INFO: Recovered parse for slot n-best match: 'all': ['todo', 'todos', 'todas', 'todos los']
INFO: src_parse: [IN:GET_ALARM [SL:AMOUNT \textbf{all} ] [SL:DATE_TIME for Friday ] ]
INFO: old_parse: [IN:GET_ALARM [SL:AMOUNT \underline{todo} ] [SL:DATE_TIME viernes ] ]
INFO: new_parse: [IN:GET_ALARM [SL:AMOUNT \underline{todas} ] [SL:DATE_TIME viernes ] ]
\end{Verbatim}
\caption{Example of Success Mode ``Slot n-best'' for \claspTS.
The sentence generated by the model uses a different word for the slot ``all'' than was
set during \textit{a priori} slot translation.
(Here, the feminine plural form ``tod\underline{as}'' instead of the masculine singular form ``tod\underline{o}''.)
Instead of discarding this example with Missing Slot failure mode, we can
use our pre-computed n-best slot mapping to recover a version of the target-language
parse which matches the words in the model's output.}
\label{fig:example_mtop_slot_nbest}
\end{figure}

\begin{figure}[h!]
\begin{Verbatim}[fontsize=\scriptsize, frame=single, commandchars=\\\{\}]
\textbf{INPUT}:
[CLM] \textcolor{red}{Semantic Parse for English:}
  [IN:CREATE_REMINDER [SL:PERSON_REMINDED me ] [SL:TODO 
    [IN:GET_TODO [SL:TODO check the weather ] [SL:DATE_TIME Friday ] 
    [SL:TODO see if the cookout 's still on ] ] ] ]
=> \textcolor{teal}{Translation in English:}
  Remind me to check the weather Friday to see if the cookout 's still on .;
\textcolor{red}{Semantic Parse for German:}
  [IN:CREATE_REMINDER [SL:PERSON_REMINDED mich ] [SL:TODO 
    [IN:GET_TODO [SL:TODO das Wetter überprüfen ] [SL:DATE_TIME am Freitag ] 
    [SL:TODO sehen ob die Grillparty noch stattfindet ] ] ] ]
=> \textcolor{teal}{Translation in German:}
  Erinnere mich am Freitag das Wetter zu überprüfen um zu sehen ob die Grillparty noch stattfindet .;
\textcolor{red}{Semantic Parse for English:}
  [IN:UPDATE_CALL [SL:CONTACT_ADDED nicole ] ]
=> \textcolor{teal}{Translation in English:}
  join nicole to this call;
\textcolor{red}{Semantic Parse for German:}

\textbf{OUTPUT}:
[IN:UPDATE_CALL [SL:CONTACT_ADDED \textbf{nicole} ] ]
=> \textcolor{teal}{Translation in German:} \underline{Nicole} zu diesem Anruf hinzufügen;

\textbf{FIX CASING RECOVERY}:
INFO: Recovered parse via Fix Casing: changed 'nicole' to 'Nicole'
INFO: old_parse: [IN:UPDATE_CALL [SL:CONTACT_ADDED \textbf{nicole} ] ]
INFO: new_parse: [IN:UPDATE_CALL [SL:CONTACT_ADDED \underline{Nicole} ] ]
\end{Verbatim}
\caption{Example of Success Mode ``Fix Casing'' for \claspTB.
The model generates both the parse and text, however the casing for the slot `Nicole'
does not match. Instead of discarding this example as Missing Slot
failure mode, we recover the correct parse by finding a case-insensitive match
for the slot in the text, and updating the parse to match.
}
\label{fig:example_mtop_fix_casing}
\end{figure}

\section{Filtering Machine Translation Outputs}
\label{sec:filtering_mt}

For mTOP Machine Translation experiments (either using Opus or the 20B LLM,
described in Section \ref{sec:baselines}),
we filter the outputs using heuristics to avoid noisy alignments.\footnote{
Early experiments showed these filtering mechanisms to provide
significant improvement over using the alignment as-is.
Future work can continue to explore cleaning and filtering methods for
MT alignment.}

\begin{figure*}[h!]
    \centering
    \includegraphics[width=0.4\textwidth]{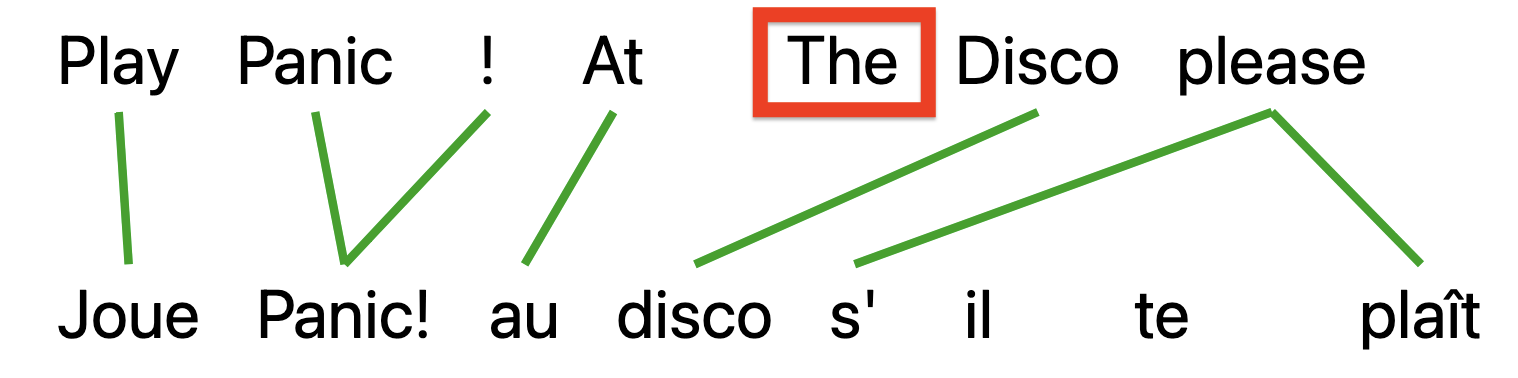}
    \caption{Example of a translation alignment discarded due to ``Missing Slot Value'',
    where a source-side slot word
    (``The'') is not aligned to any output word. The parse for the English utterance is
    \texttt{[IN:PLAY\_MUSIC [SL:MUSIC\_ARTIST\_NAME Panic ! At \underline{The} Disco ] ]}.
    (Via https://simalign.cis.lmu.de/)}
    \label{fig:example_missing_slot}
\end{figure*}

\begin{figure*}[h!]
    \centering
    \includegraphics[width=0.25\textwidth]{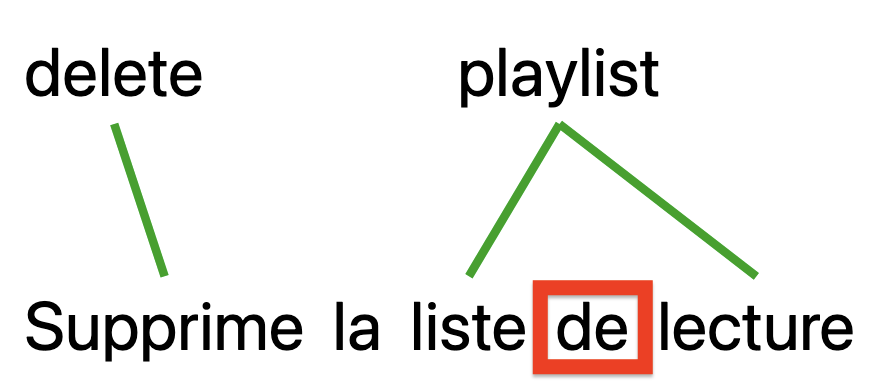}
    \caption{Example of a translation alignment discarded due to ``Discontiguous Target'',
    where a source-side slot (``playlist'') aligns to a discontiguous set of words in the target (``liste'' and ``lecture'', missing ``de'').
    The parse for the English utterance is
    \texttt{[IN:DELETE\_PLAYLIST\_MUSIC [SL:MUSIC\_TYPE \underline{playlist} ] ]}.
    (Via https://simalign.cis.lmu.de/)}
    \label{fig:example_discont}
\end{figure*}

\begin{figure*}[h!]
    \centering
    \includegraphics[width=0.7\textwidth]{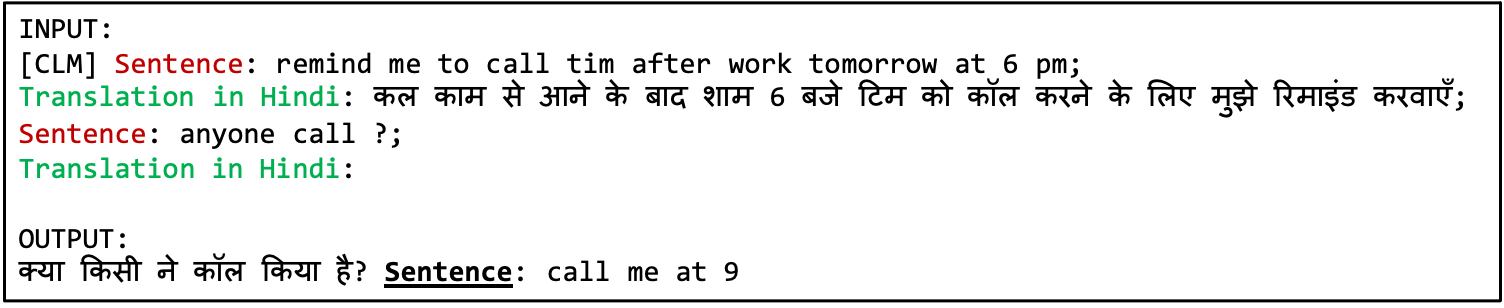}
    \caption{Example of a translation output from the 20B model, discarded due to Contains ``Sentence''.}
    \label{fig:example_contains_sentence}
\end{figure*}

We first apply Sim-Align \citep{jalili-sabet-etal-2020-simalign}
to align the translated sentence back to the original English source, in order to
compute the parse in the target language.
We discard outputs which exhibit any of four Failure Modes.
The first two Failure Modes are related to slots:
(i) Missing Slot Value
(Figure \ref{fig:example_missing_slot}); or
(ii) Discontiguous Target (Figure \ref{fig:example_discont}).
We also discard outputs which:
(iii) Copy the Original input text verbatim,
and in the case of translation with the 20B model,
(iv) contain the word "Sentence", i.e. fail to end with a semicolon as prompted
(Figure \ref{fig:example_contains_sentence}).

\textbf{We define the ``Success Rate'' as the percentage of remaining outputs after filtering.} As show in Table \ref{tab:mt_success_rate},
the success rate is far from 100\%, e.g. for Opus MT varying from
86.7 for German (``de'') down to 62.2 for Hindi (``hi'').
This reflects the \textit{difficulty of the alignment task,
a fundamental limitation of the baseline approach of Machine Translation with slot alignment},
particularly between distant language pairs such as English and Hindi.

Also of note, when using the 20B model for translation, 13.4\% of the prompts for Hindi were
discarded due to producing the word "Sentence", i.e. not
ending with a semicolon as instructed.
(See an example in Figure \ref{fig:example_contains_sentence}, compared to
Figure \ref{fig:example_in_context_text_mt}.)
We hypothesize this could be caused by using
a semicolon as the separator, which might be less common in Hindi than the
other languages which use the Latin alphabet.
Future work could explore using language-agnostic separators such as
\texttt{<br>}.

\begin{table}[h!]
\footnotesize
\centering
\begin{tabular}{c|cc||ccc|c}
\hline
\multirow{3}{*}{\makecell{MT\\Model}} &
\multirow{3}{*}{Language} &
\multirow{3}{*}{\makecell{\textbf{Success}\\\textbf{Rate}}} &
\multicolumn{4}{c}{Failure Modes} \\\cline{4-7}
  &    &    & \makecell{Missing\\Slot Value}   & \makecell{Discontiguous\\Target}   & \makecell{Copy\\ Original}  & \makecell{Contains\\``Sentence''} \\
\hline
 \multirow{5}{*}{Opus}    & de         &           86.7 &            4.6 &             8.6 &             0.1 &                --   \\
 & es         &           74.2 &            4.9 &            20.8 &             0.2 &                --   \\
 & fr         &           82.3 &            3.9 &            13.7 &             0.1 &                --   \\
 & hi         &           \underline{62.2} &           19.4 &            18.4 &             0.0 &                --   \\\cline{2-3}
 & avg        &           \textbf{76.4} &           --   &            --   &            --   &                --   \\
 \hline
 \multirow{5}{*}{20B}     & de         &           85.5 &            4.2 &             9.8 &             0.3 &                 0.1 \\
 & es         &           70.9 &            6.6 &            21.5 &             0.6 &                 0.3 \\
 & fr         &           77.4 &            4.7 &            17.2 &             0.1 &                 0.5 \\
 & hi         &           \underline{58.3} &           12.2 &            16.0 &             0.1 &                13.4 \\\cline{2-3}
 & avg        &           \textbf{73.0} &           --   &            --   &            --   &                --   \\
\hline
\end{tabular}
\caption{Success Rate (percentage) and occurrence of failure cases (percentage) of 
Machine Translation (MT) with with alignment across MT models and languages.
The average across the four languages is bolded, and
the language with lowest (i.e., worst) Success Rate for each model is underlined.}
\label{tab:mt_success_rate}
\end{table}

\section{Sim-Align Settings}
\label{sec:simalign_settings}
We explore four settings for Sim-Align, using either (multilingual)
``bert'' \citep{devlin-etal-2019-bert} or ``xlm-roberta-base'' 
\citep{conneau-etal-2020-unsupervised}
each with either ``ArgMax'' or ``IterMax'' as the alignment method.
\textit{We choose ``bert'' with with ``IterMax''} as we find it has the highest
Success Rate (defined in Appendix \ref{sec:filtering_mt}).

\end{document}